\documentclass{article}

\usepackage[final]{neurips_2021}

\usepackage[utf8]{inputenc} 
\usepackage[T1]{fontenc}    
\usepackage{hyperref}       
\usepackage{url}            
\usepackage{booktabs}       
\usepackage{amsfonts}       
\usepackage{nicefrac}       
\usepackage{comment}       
\usepackage{microtype}      
\usepackage{xcolor}         
\usepackage{graphicx}
\usepackage{minitoc}
\usepackage{subcaption}
\usepackage{fancyvrb}
\usepackage{pythonhighlight}
\usepackage{pdfpages}
\usepackage{subcaption}
\usepackage{multirow}

\usepackage{xpatch}
\makeatletter
\xpatchcmd{\sv@part}{\huge \bfseries \partname \nobreakspace \thepart \par \vskip 20\p@ \fi \Huge \bfseries #2}{\fi \Huge \bfseries \thepart. #2}{}{}
\makeatother

\usepackage{natbib}
\usepackage{apalike}
\usepackage{array}
\newcolumntype{L}{>{\left\arraybackslash}m{15cm}}

\title{\Large Recursively Summarizing Books with Human Feedback}

\author{%
  Jeff Wu\thanks{This was a joint project of the OpenAI Alignment team.  JW and LO contributed equally.  DMZ, NS, and RL were full-time contributors for most of the duration.  JL and PC managed the team.  Corresponding author jeffwu@openai.com.
  }
  \And
  Long Ouyang$^*$ 
  \And 
  Daniel M. Ziegler$^*$
  \And 
  Nisan Stiennon$^*$  
  \And
  Ryan Lowe$^*$
  \And 
  Jan Leike$^*$
  \And 
  Paul Christiano$^*$
  \AND
  \normalfont{OpenAI}
}

\begin{document}

\maketitle

\doparttoc 
\faketableofcontents 

\begin{abstract}
A major challenge for scaling machine learning is training models to perform tasks that are very difficult or time-consuming for humans to evaluate. We present progress on this problem on the task of abstractive summarization of entire fiction novels. Our method combines learning from human feedback with recursive task decomposition: 
we use models trained on smaller parts of the task to assist humans in giving feedback on the broader task. 
We collect a large volume of demonstrations and comparisons from human labelers, and fine-tune GPT-3 using behavioral cloning and reward modeling to do summarization recursively.
At inference time, the model first summarizes small sections of the book and then recursively summarizes these summaries to produce a summary of the entire book. Our human labelers are able to supervise and evaluate the models quickly, despite not having read the entire books themselves.
Our resulting model generates sensible summaries of entire books, even matching the quality of human-written summaries in a few cases ($\sim5\%$ of books).
We achieve state-of-the-art results on the recent BookSum dataset for book-length summarization.
A zero-shot question-answering model using these summaries achieves competitive results on the challenging NarrativeQA benchmark for answering questions about books and movie scripts.  We release datasets of samples from our model.\footnote{See \href{https://openaipublic.blob.core.windows.net/recursive-book-summ/website/index.html}{https://openaipublic.blob.core.windows.net/recursive-book-summ/website/index.html}}

\end{abstract}

\section{Introduction}
\label{sec:intro}

To train an ML model on a new task, we need a training signal that tells the model which behaviors are better and which are worse. For some tasks, like playing a video game, this training signal can be calculated automatically. 
However, for many useful tasks an accurate training signal can only be provided via a human in the loop. For example, humans can provide demonstrations of the correct behavior \citep{bain1995framework} or compare two outputs from the model being trained \citep{christiano2017deep}, and this data is used to train the model. 

In this paper we focus on tasks that are difficult for humans to supervise or evaluate, either because the tasks take a lot of time or because they require specialized knowledge and expertise to evaluate.  For example, imagine training a model to summarize an entire sub-field of scientific research.  For a human to provide a demonstration or evaluate the quality of a model-generated summary, they would likely need a huge amount of time and expertise. One could circumvent this difficulty by using easier-to-measure proxy objectives (e.g. how often words in the summary relate to the topic, and how accurate individual sentences in the summary are), but these proxies are usually less aligned with our actual goals, and optimizing them can have unintended consequences~\citep{clark2016faulty, krakovna2020specification, amodei2016concrete}. 

\begin{table}[]
    \centering
    \small
    \begin{tabular}{>{\arraybackslash}m{14cm}}
    \toprule 
    \textbf{Summaries of ``A Promised Land'' by Barack Obama}\\ \hline
        \multicolumn{1}{c}{\textbf{Best human-written  summary}} \\
        Barack Obama details his journey from a flippant Hawaiian high schooler raised by a traveling, unconventional mother to the first African American President of the United States. His political career is influenced by the tales of his mother, grandmother Toot, his childhood international travels, and civil rights leaders. While he’s in Harvard Law, he meets and marries Michelle Robinson. They have two daughters, Malia in 1998 and Sasha in 2001, while he is a Democratic Illinois Senator. Barack juggles marriage and fatherhood with a political life that keeps him from home. This often causes strain with Michelle. In 2004, he wins a seat in the U.S. Senate. By 2006, several Democratic senators urge him to run for President in 2008.  He feels a good campaign could shake loose the traces of America’s racial past. He wins the primary after a contentious battle with Hillary Clinton. After a groundbreaking grassroots campaign, Barack wins the general election against Republican John McCain. [...]
        \\ \hline
        
        \multicolumn{1}{c}{\textbf{Best 175B RL summary}} \\ 
        Barack Obama wins the presidency in 2008, and his first term is marked by economic turmoil and political polarization. His domestic policy includes the Affordable Care Act, which reformed healthcare in the United States.

Obama's presidency includes terrorism, the Arab-Israeli conflict, and the Deepwater Horizon oil spill. His foreign policy includes finding and eliminating Osama bin Laden. After the successful Abbottabad raid, Obama reflects on the unity of effort displayed by all the individuals involved in the operation. He wonders if that same unity of effort could be applied to other areas of government.
        \\ \hline 
       \multicolumn{1}{c}{\textbf{Best 175B BC summary}}\\  
        Barack Obama writes a book about his experiences as president. He explains that he started questioning the U.S. government when he was at Occidental College. He didn't believe in politics, but was inspired by social movements. He ran Project VOTE! in advance of the 1992 presidential race. Obama thinks that the politics are going to be brutal. He introduces Laura. She has been diagnosed with aggressive breast cancer that had spread to her bones. The U.S. policy in the Middle East focuses narrowly on maintaining stability, preventing disruptions to our oil supplies, and keeping adversarial powers from expanding their influence.
        \\
    \bottomrule
    \end{tabular}
    \caption{Summaries of ``A Promised Land'' written by our models (175B RL policy and 175B behavior cloning policy) and a human. Since this book was written in 2020, it was not seen by our model during either pretraining or fine-tuning. For each model, we select the summary with the highest Likert rating according to human evaluations (out of 6 summaries for RL, 9 for BC, and 2 for the human-written summary). See \href{https://openaipublic.blob.core.windows.net/recursive-book-summ/website/index.html}{our website} for all human and model summaries. }
    \label{tab:main-ex}
\end{table}

Successfully training ML systems on such tasks will require more scalable means of producing an effective training signal --- this problem is known as \textit{scalable oversight} \citep{amodei2016concrete}. 

Our approach to scalable oversight is directly inspired by \cite{christiano2018supervising} and \cite{leike2018scalable}, who make use of \textit{task decomposition} \citep{singh1992transfer,dayan1993feudal} and learning from human feedback.
At a high level, these methods take a top-level task and decompose it into several smaller subtasks whose answers would help a human solve or evaluate the top-level task. These subtasks can in turn be decomposed into smaller tasks until it is feasible for humans to provide a training signal for a leaf task. ML models can be trained to solve the leaf tasks, to solve higher-level tasks given answers to the lower-level tasks, and to decompose the harder tasks into subtasks. While \cite{dayan1993feudal} and \cite{christiano2018supervising} only tried this on simple algorithmic tasks, \cite{perez2020unsupervised} and \cite{rajani2019explain} use similar ideas for question-answering tasks using a single step of decomposition.

We take a step further in this direction by scaling task decomposition to abstractive book summarization.  Abstractive book summarization is a difficult task, where dataset collection is challenging~\citep{mihalcea2007explorations,ladhak2020exploring,kryscinski2021booksum} and existing methods are typically either extractive~\citep{radev2004mead,mihalcea2004textrank,bamman2013new,ladhak2020exploring} or focused on shorter stories~\citep{kazantseva2006approach,zhang2019generating}.

We implement a natural task decomposition for long-form summarization: first, we train models to summarize small parts of the book, and then use these models to help humans summarize larger sections of the book, and continue with this strategy recursively.  We train a single model to perform these tasks using standard cross-entropy behavioral cloning (BC) and reinforcement learning (RL) from human preferences \citep{christiano2017deep}.  

Our main result is a model that can be applied recursively to generate plausible summaries of entire books.  Our approach lets us summarize books of arbitrary length -- we achieve believable summaries on books with hundreds of thousands of words by recursing to depth 3.  With a non-recursive approach, generating or evaluating a book summary requires a human reading the entire book, so naively collecting such a dataset is over 50x more expensive per data point (see Appendix \ref{sec:humantime_estimates}).

\begin{figure}
    \centering
    \includegraphics[width=1.\linewidth, clip, trim=0mm 90mm 0mm 50mm]{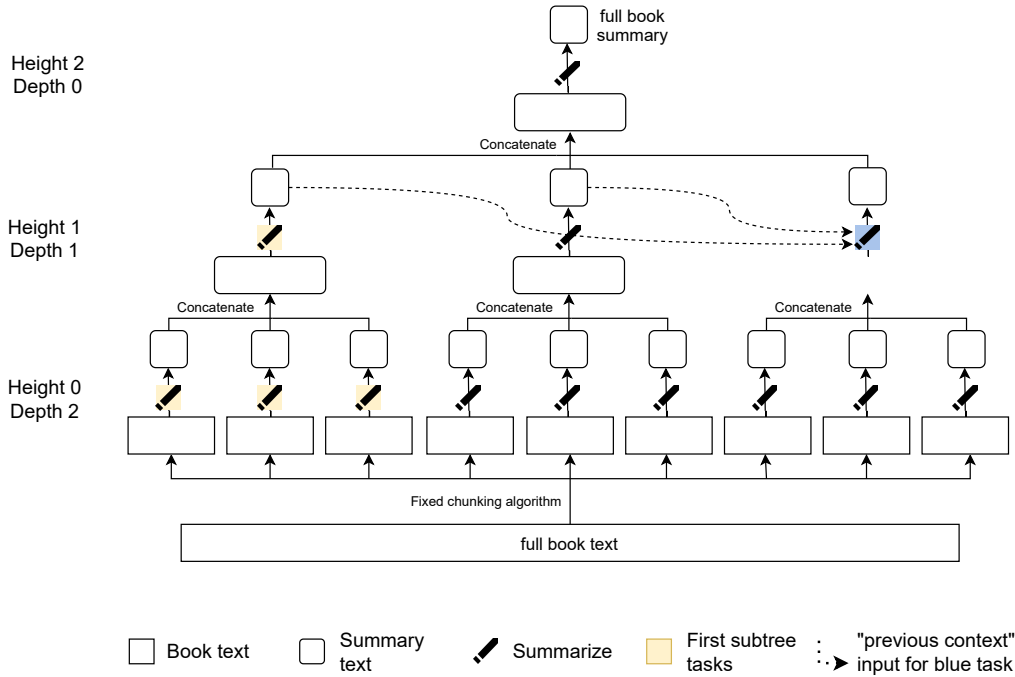}
    \caption{\label{fig:tree}
    Our procedure for summarizing books that combines task decomposition with learning from human feedback. We first decompose the book text into multiple chunks using a fixed (not learned) chunking algorithm (height 0). We then collect demonstrations from humans summarizing these chunks, and train an ML model on this data using behavior cloning (each node with a pencil symbol corresponds to a summarization task carried out by either a human or model). We can then collect human data comparing different model outputs, and use this data to further train the summarization policy using reward modeling~\citep{christiano2017deep}. We then concatenate several height 0 summaries, collect data for summarizing these summaries, and fine-tune our model on this summarization task (height 1). We repeat this procedure recursively until we've summarized the entire book. We use the same policy to summarize text at all levels. Summarization tasks later in the book are conditioned on previous summaries at the same height; we show this for the blue task at height 1 using dotted arrows, but this happens at all levels of the tree.
    \vspace{-5mm}}
\end{figure}

Qualitatively, these summaries contain important events from the book, and sometimes synthesize these details abstractively; however, they often leave out important details or fail to grasp the broader context. When evaluated quantitatively, our model significantly outperforms our behavioral cloning baseline, and a small number of summaries approach human-level quality. Separately, we perform an ablation comparing RL to BC on summarizing smaller sections of a book, and find that RL has better scaling properties.
We also evaluate our summaries with the NarrativeQA question-answering dataset \citep{kovcisky2018narrativeqa} and find that a zero-shot model taking our summaries as input achieves competitive results at answering questions about books and movie scripts.  We also achieve state-of-the-art results on the recent BookSum dataset \citep{kryscinski2021booksum} for book-length summarization.

Overall, our results show that combining recursive task decomposition with learning from human feedback can be a practical approach to scalable oversight for difficult long-document NLP tasks.  We hope that our work encourages more research in using models trained on simpler tasks to aid humans in providing training signals on more difficult tasks.

\section{Approach}
\label{sec:approach}

\subsection{Task decomposition}

Consider a task for which it is very expensive for a human to provide a training signal. \cite{christiano2018supervising}, \cite{irving2018ai}, and \cite{leike2018scalable} all propose in some way reducing the task into simpler parts which humans can supervise.

In task decomposition, a human decomposes this parent task into several subtasks, such that each subtask is simpler than the parent task, and having the responses to the subtasks would help a human provide a training signal for the parent task.  This task decomposition process can be applied recursively to obtain a tree of tasks, such that the leaf tasks are simple enough for a human to demonstrate or evaluate. For example, the parent task “Write a research report on climate change interventions” might decompose into a subtask like: “Give me a list of the most promising climate change interventions”, which then further decomposes into simpler tasks like “How effective is reducing food waste?” and “What are ways to make nations coordinate in avoiding tragedy of the commons scenarios?”.

If we repeat this process many times, we obtain a dataset that we can use to train an ML model. Specifically, given a (sub)task we want to train a model that can perform two fundamental operations:
\begin{enumerate}
\item{Decompose: Ask for responses to a set of simpler tasks.  }
\item{Respond: Given responses to some number (possibly none) of simpler tasks, respond to the original task.  When there are simpler tasks used, we sometimes refer to the operation as Compose, since it composes the sub-responses into an overall response.}
\end{enumerate}

Then any task can be performed via a recursive procedure if it is amenable to decomposition; we show a pseudocode implementation in Appendix \ref{sec:decompdetails}.  It remains an open question to what extent natural tasks are actually amenable to decomposition \citep{evaluating2020ought}.  

While the framework above is fully general, it can be further simplified if the task lends itself to a simple recursive structure where the decomposition operation can be performed algorithmically, and the ML model only needs to be trained on the Respond operation.

\subsection{Decomposition for book summarization}
\label{sec:booktree}

We use a simple procedure to algorithmically decompose a summarization task for a piece of text:  If the text is short enough, summarize it directly.  If it is longer, chunk the text into smaller pieces, and recursively ask to summarize each one.  This results in a tree of summarization tasks (see Figure \ref{fig:tree}), where only the leaf tasks operate on passages of the original book text.  

Each task, corresponding to nodes with pencil symbols in Figure \ref{fig:tree}, has a height and depth, which correspond to the standard terminology used for trees.  
The \textbf{height} of a node is the length of the longest downward path to a leaf from that node.  A height 0 task is a leaf task, where the goal is to summarize the original book text.  We sometimes refer to tasks that are height > 0 as composition tasks, since the input is a concatenation of summaries, and the goal is to produce another summary.   
The \textbf{depth} of a node is the length of the path from the node to the root.  A depth 0 task is the final summarization task, where the goal is to produce a summary of an entire book (given summaries produced from the depth 1 tasks).

An evident issue with the above approach is that tasks corresponding to passages further into a book may lack the necessary context for a successful summary.  We remedy this by additionally putting prior summaries in context, from the same depth, concatenated together in order.\footnote{Early on, we found this previous context to help the model (according to log loss on a BC model).  We also found that variants that include the previous un-summarized text did worse -- though it includes more information, our models did not have enough context length to make use of it.}  We call these summaries the \textbf{previous context}.  In Figure \ref{fig:tree}, the previous summaries inputs for the blue task are indicated using dotted lines.  We include as many prior summaries as can fit in the model's context length.  We would like each summary to flow naturally from the previous context, since it may get concatenated with it at a higher height or in the previous context for a later task.

A convenient property of this decomposition is that all of the tasks in the tree are extremely similar to one another.  Every task for the model is a summarization task that can be formatted the same way.  The input text is either the original book text or a concatenation of summaries, and we optionally have additional previous context in the form of summaries.  

Pseudocode and detailed parameters of tree construction can be found in Appendix \ref{sec:pseudocode_book}.

\subsection{Training}
\label{sec:training}

For training the model, we most closely follow the procedure of \cite{stiennon2020learning}.  We start with a pretrained language model and a pool of trained human labelers (see Appendix \ref{sec:labelers} for details).  We collect demonstrations from labelers and train a model via behavioral cloning.  We then repeat many iterations of reward learning and reinforcement learning.  To learn the reward function, we collect comparisons from labelers on outputs from the current best policy and train a reward model to predict log odds that a response is better.  Reinforcement learning directly optimizes the reward with an additional KL term to prevent too much drift from the initial policy, typically our best supervised policy.  More details in Appendix \ref{sec:finetuning_hyperparameters}.

To collect a label for a given task, we need to generate its inputs: if a node is not a leaf, we run an existing model (typically the best available) recursively to generate summaries for each of its children.

In summary, we use the following algorithm:
\begin{enumerate}
\item{Recursively decompose books (and compose child summaries) into tasks using the procedure described in \ref{sec:booktree}, using the best models we have\footnote{While the tree is typically created from a single best model for all tasks, there are times when, e.g., our best model at height 0 is an RL model but the best model at height 1 is supervised.  We also initially experimented with training different models for height 0 and height 1, but found that training a unified model worked better, and trained a single model for all heights thereafter.  } and the best sampling parameters we have\footnote{Our best guess sampling parameters are generally determined by human evaluations on the individual tasks.  See Appendix \ref{sec:temperature}}.  While this could be done with humans, it would be prohibitively expensive.}
\item{Sample a node from the tree, corresponding to a summarization task which we’d like to train on.\footnote{Note that throughout much of the project, we sample only from the early parts of the tree and thus avoid running the full procedure from step 1.}  Details below in \ref{sec:curriculum}.}
\item{Obtain training data, given the inputs to that node
\begin{enumerate}
    \item For demonstrations, we then have human labelers write a desired output
    \item For comparisons, we run the model we wish to train to obtain two outputs, typically at temperature 1.  We then ask human labelers to choose which output is better.
\end{enumerate}
}
\item{We then finetune the model using the training data \begin{enumerate}
    \item For demonstrations, we use behavior cloning (BC).  We do a supervised finetune using the standard cross entropy loss function.
    \item For comparisons, we use reinforcement learning (RL) against a reward model trained to predict human preferences.  
    \end{enumerate}
}
\end{enumerate}

We can iterate this entire process with newer models, different node sampling strategies, and different choice of training data type (demonstration versus comparison).

\subsubsection{Auto-induced distributional shift}
\label{sec:ads}

Since each model is trained on inputs produced by a different model, inputs produced by itself are outside of the training distribution, thus causing auto-induced distributional shift (ADS) \citep{krueger2020hidden}.  This effect is more severe at later parts in the tree computation (later in the book, and especially higher in the tree).  This means that after each round of training, running the full procedure always results in inputs out of the prior training distributions, for tasks at non-zero height.  While we did not systematically measure the severity of this effect, in practice we generally found that additional rounds of training at height 0 resulted in better-rated summaries at height 1.

\subsubsection{Training curriculum}
\label{sec:curriculum}

Because of the ADS mentioned in Section \ref{sec:ads}, it is advantageous to prioritize training on nodes earlier/lower in the tree computation, before moving to nodes later in the computation.  

We define the following terms:
\begin{itemize}
\item \textbf{First subtree}.  The first subtree refers to the first height 1 task, and its height 0 child tasks (of which there are typically 10-13).  See the yellow nodes in Figure \ref{fig:tree} for an example.
In Section \ref{sec:fullbookevals}, we find that by training on merely the first subtree, the model can generalize to the entire tree.
\item \textbf{First leaves}.  The first leaves refers to the height 0 tasks in the first subtree, i.e. those which are children of the first height 1 task.
\end{itemize}

For early rounds, we initially train only on the first leaves, since inputs to later nodes depend on having plausible summaries from earlier nodes, and we do not want to use excessive human time.  We then move to the entire first subtree (additionally training on a single height 1 task), once the summaries for the first leaves look reasonable.  At this point, our model is already capable of generalizing to the full tree, and we switch to training on all nodes. Curriculum changes were made in an ad hoc manner, moving on when we deemed the models "good enough" at earlier tasks.

\subsubsection{Fine-tuning details}
\label{sec:finetune}

We use pretrained transformer language models \citep{vaswani2017attention} from the GPT-3 family \citep{brown2020language}, which take 2048 tokens of context.  Input tokens are produced by the byte pair encoding introduced in \cite{radford2019language}.  Other architecture and hyperparameters choices follow those of \cite{stiennon2020learning}.  More details in Appendix \ref{sec:finetuning_hyperparameters}.

\paragraph{Behavioral cloning and reward modeling}
In the \textit{first leaves} phase of the project, we collect data for all first leaves together.  When moving to \textit{first subtree}, we independently collect data for the height 1 tasks, letting us vary the ratio of training data at the different heights.  Finally, for the \textit{full tree} phase, we follow a strategy of first randomly sampling a depth, and then randomly selecting a task amongst tasks at that depth.  Inputs are typically generated using the best model available and best guess sampling parameters (see Appendix \ref{sec:temperature}).

In all cases, we train on all past data (individual demonstrations and comparisons for tasks from various parts of the tree).  We then shuffle and sample tasks randomly.  
\paragraph{Reinforcement learning}
\label{sec:rl_variants}
We ran three variants of sampling tasks for reinforcement learning episodes, corresponding to our changes in the training curriculum.

\begin{enumerate}
\item The first leaves: Each episode is a single first leaf task.  The algorithm trains on consecutive leaf tasks in succession; the sampled summaries are used as previous context for later leaves.
\item The first subtree:  Each episode consists of a first leaf task or the height 1 composition task for the first subtree.  The algorithm trains on the leaf tasks in succession, followed by the composition task using their sampled outputs.
\item Full tree:  We choose a random depth $d$ and then a random node at that depth.  The algorithm trains on N successive depth $d+1$ tasks followed by a single depth $d$ composition task using those N outputs.  Input trees are generated ahead of time from the initial model with best-effort sampling settings (in practice, we sometimes use some trees from older models as well). 
\end{enumerate}

Since our demonstration and comparison data is at the level of individual nodes, we train the RL policy at the same granularity: each task is its own episode, and no rewards propagate to other nodes of the tree.  

\subsection{Advantages of decomposition}

Compared to end-to-end training, decomposition makes it much easier to collect human feedback for a given task.  Correspondingly, it makes the task much easier for the ML model.  But it also offers other benefits:
\begin{enumerate}
\item It empowers a human to do or evaluate parts of the task themself. For example, a human with access to lower-level summaries can quickly summarize themselves.  
\item It makes it easier to trace what the model is thinking, and debug errors in the model. If a model summary contains a relatively isolated fact, a human with access to the tree can trace it back to the original text.
\item Our procedure generalizes gracefully to longer books.  It can be used at test time on books of unbounded length, regardless of the length of books in the training dataset.

\end{enumerate}

\section{Task details}

\subsection{Training dataset}
For training, we use a subset of the books used in GPT-3's training data~\citep{brown2020language}. The books are primarily fiction, and contain over 100K words on average.  We further constrain our dataset by asking labelers to skip non-narrative books.

We chose narrative fiction books due to our belief that they were the most difficult to summarize, which is supported by our later qualitative findings (Appendix \ref{sec:qualitative}).
Summarizing narrative texts is particularly challenging for extractive methods since any given sentence tends to be a very low-level description.  
We find additional evidence for this in Section \ref{sec:booksum}, where our models outperform an extractive oracle on the BERTScore metric.

\subsection{Summarization task}

We aim to summarize abstractively, tracing out narrative arcs and larger themes rather than listing series of events. Our primary metric is labeler judgments of overall summary quality on a 1-7 Likert scale, on held-out books that were neither in the GPT-3 pretraining dataset nor in our book dataset. 
We also ask labelers to evaluate summary accuracy, coverage of the source text, coherence, and amount of abstraction; see more details on our instructions to labelers in Appendix \ref{sec:labeler_guidelines}.

For each summarization subtask, we generally aim to compress the text by a factor of 5-10x, with length upper limits of 128 to 384 tokens, depending on the task height.  We ask labelers to evaluate summary quality conditioned on its length; that is, labelers are answering the question ``how good is this summary, given that it is X words long?'' This is in part to avoid the scenario where, if longer summaries are preferred by labelers, models will generate the longest summaries allowed by the length constraints \citep{stiennon2020learning}.

We emphasize that for each subtask, labelers only consider the quality of the summary with respect to the direct input to the model, rather than the subset of the book representing the true summarization target.  See Appendix \ref{sec:true_task} for more discussion.

\section{Results}

\subsection{Full book human evaluations}
\label{sec:fullbookevals}

\begin{figure}
\centering
\begin{subfigure}{.54\linewidth}
  \centering
  \includegraphics[width=1.0\linewidth]{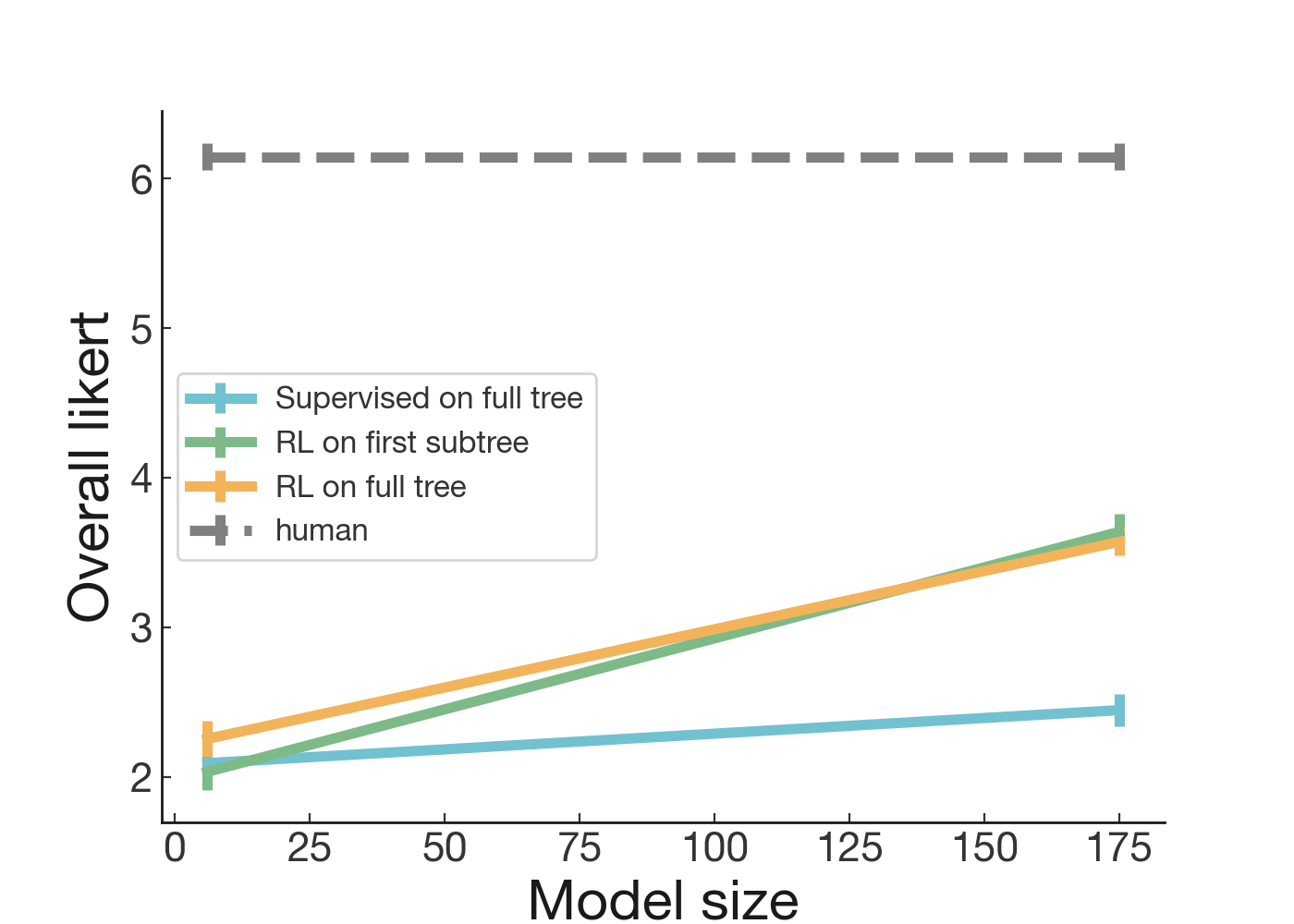}
  \caption{}
  \label{fig:fullbook_main}
\end{subfigure}%
\begin{subfigure}{.48\linewidth}
  \centering
  \includegraphics[width=1.0\linewidth]{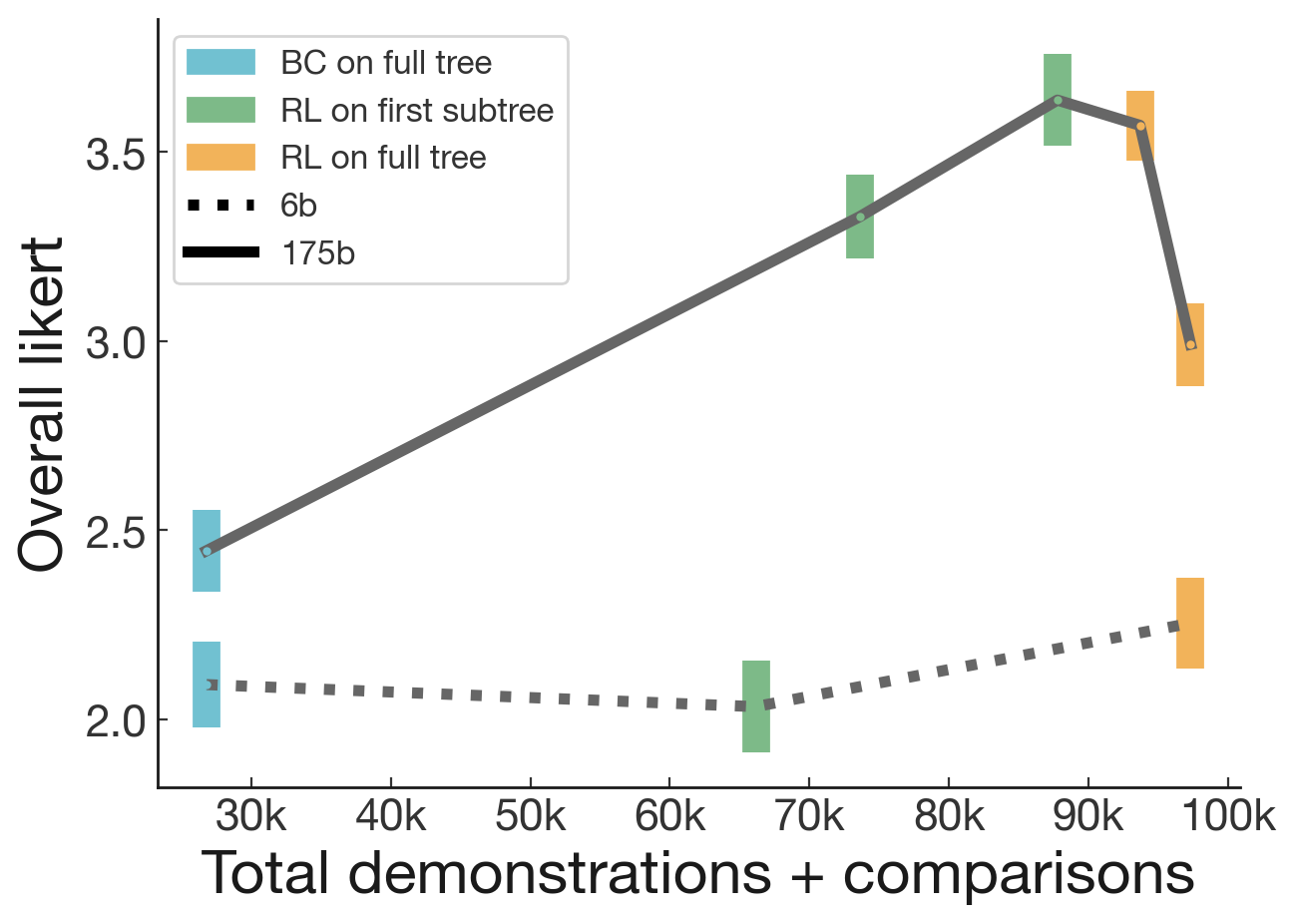}
  \caption{}
  \label{fig:fullbook_human_time}
\end{subfigure}
\caption{Results on full book evaluations, (a) as a function of model size (measured in billions of parameters), and (b) as a function of number of labels.  Error bars are obtained by averaging ratings for each book, then computing the standard error of the mean across books. At larger model sizes, our RL models significantly outperform our BC models (`supervised on full tree'). Our models are still far from human performance.}
\label{fig:fullbook}
\end{figure}

\subsubsection{Methodology}
We first evaluate our models' ability to summarize full books that were unseen during pretraining or fine-tuning. 
To do this, we use the 40 most popular books published in 2020 \href{https://www.goodreads.com/book/popular_by_date/2020}{according to Goodreads} at the time we looked.
The resulting books span a variety of genres (see Table \ref{tab:books}).  

We then assigned two labelers to read each book (purchased with reimbursement) and to write a summary of the book.  Finally, we ask the labelers to rate summaries from various models and from the other labeler.  Labeler agreement for relative quality of model-written summaries was nearly 80\%.

We evaluate two model sizes, 175B parameters and 6B parameters.  For each size, we also evaluate three different modes of training:  RL on the whole tree, RL on the first subtree, and BC on the whole tree.
For each policy, we generate 3 summaries each, in order to reduce error bars.  Even for temperature 0 policies, we can vary the summaries by changing the seed used to randomly choose chunking boundaries -- we found this to produce significant variation in the summaries.  

We evaluated all BC policies at temperatures T=0.0, 0.3, and 0.6 on this test set.  
The results in Figures \ref{fig:fullbook} and \ref{fig:fullbook_likert_hists} use the best temperatures for these policies.\footnote{While this may overstate quality of the BC policies, we consider the policies to be a baseline and did not want to understate the quality.} This is because it was too expensive to ablate temperature on the full book summarization task on our validation set (though we we show temperature sweeps on the validation set for leaf summarization tasks in Appendix \ref{sec:temperature}, these temperatures are not a priori the best for full book summarization). In the end, we empirically found that the best temperatures for the leaf task were also the best for full book summarization: T=0.6 was best for our 6B BC baseline, and all temperatures performed about equally for our 175B BC baseline.

\subsubsection{Findings}
\label{sec:fullbook_findings}

\begin{figure}
    \centering
    \includegraphics[width=0.4\linewidth]{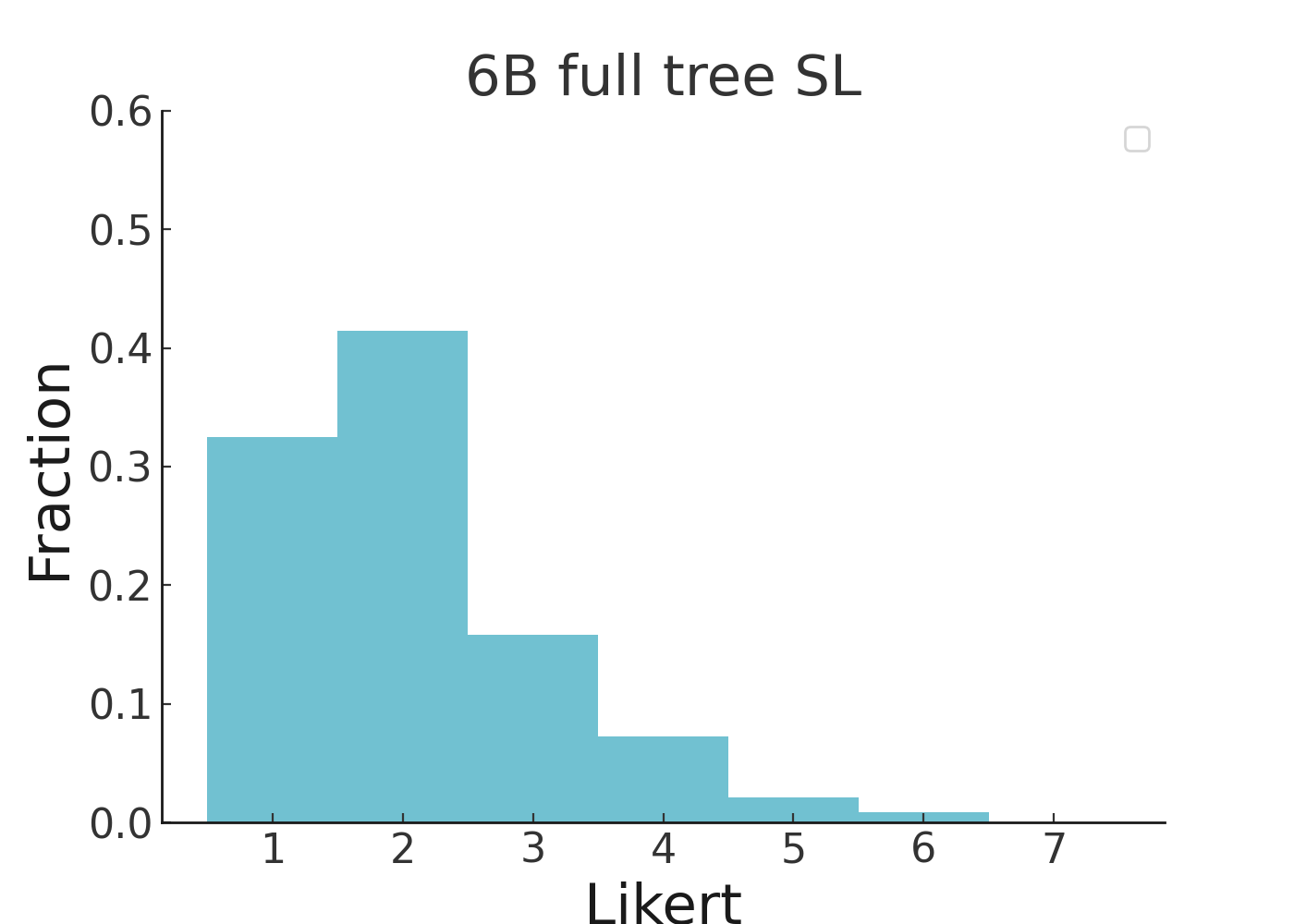}
    \includegraphics[width=0.4\linewidth]{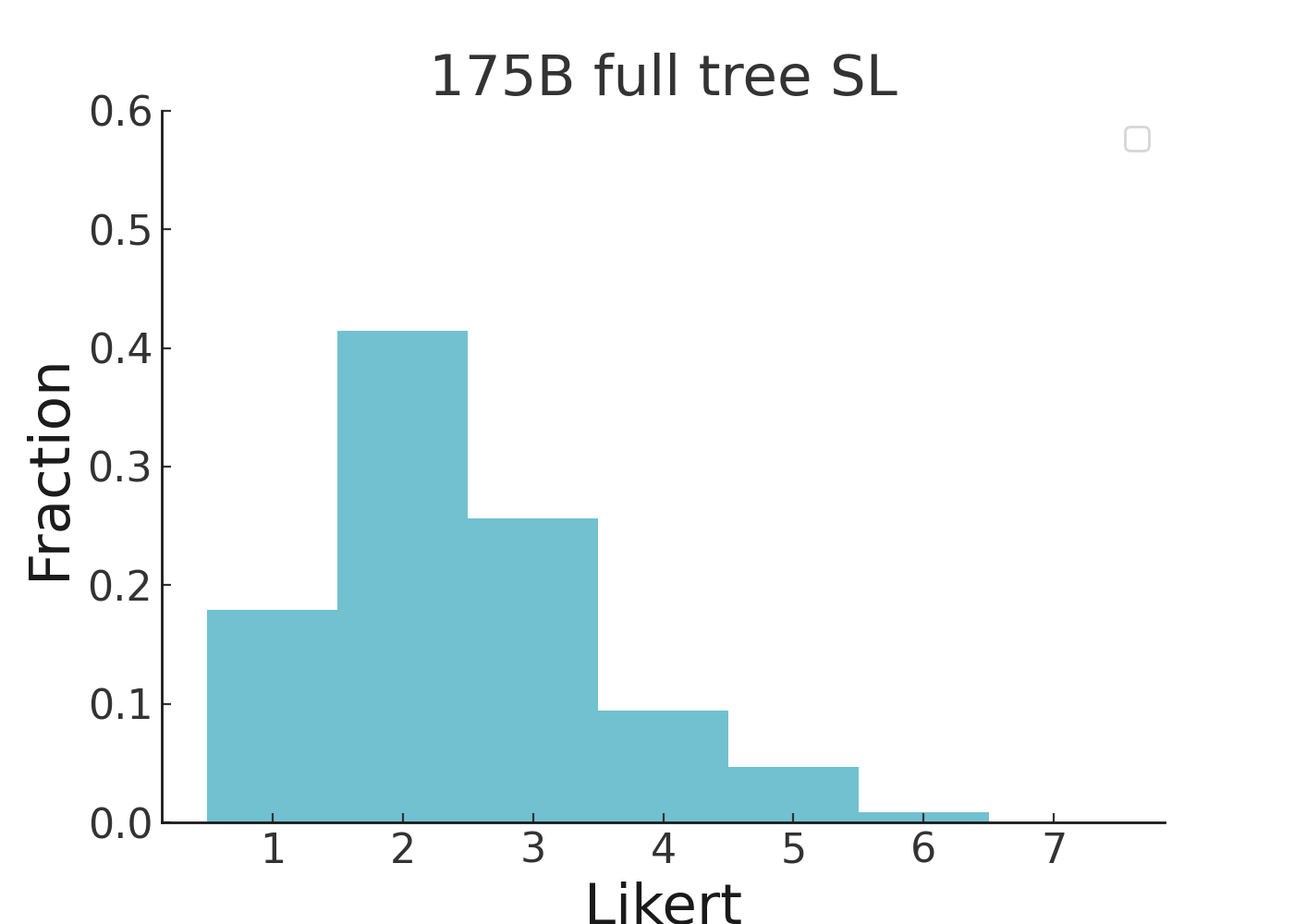}
    \newline
    \centering
    \includegraphics[width=0.4\linewidth]{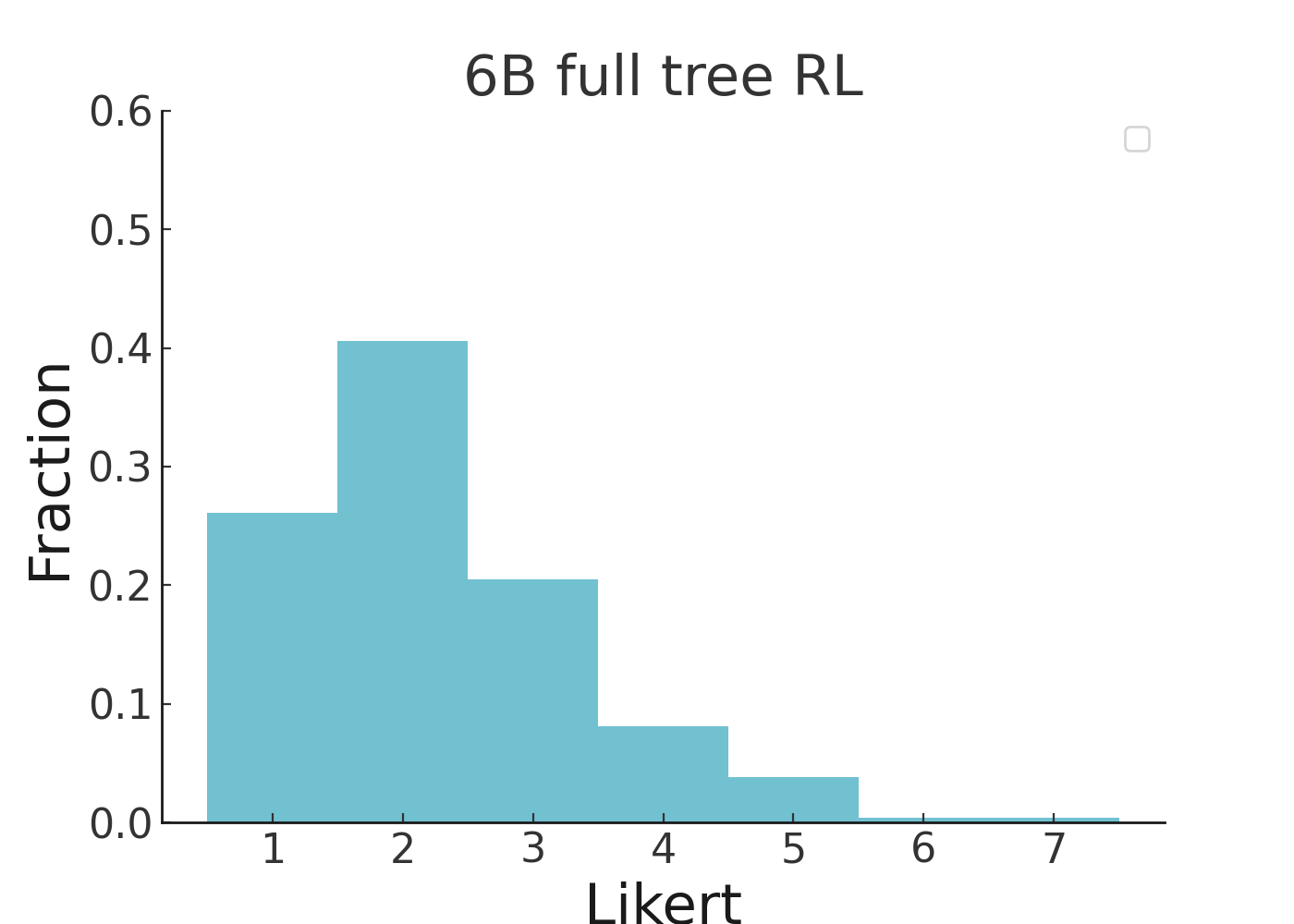}
    \includegraphics[width=0.4\linewidth]{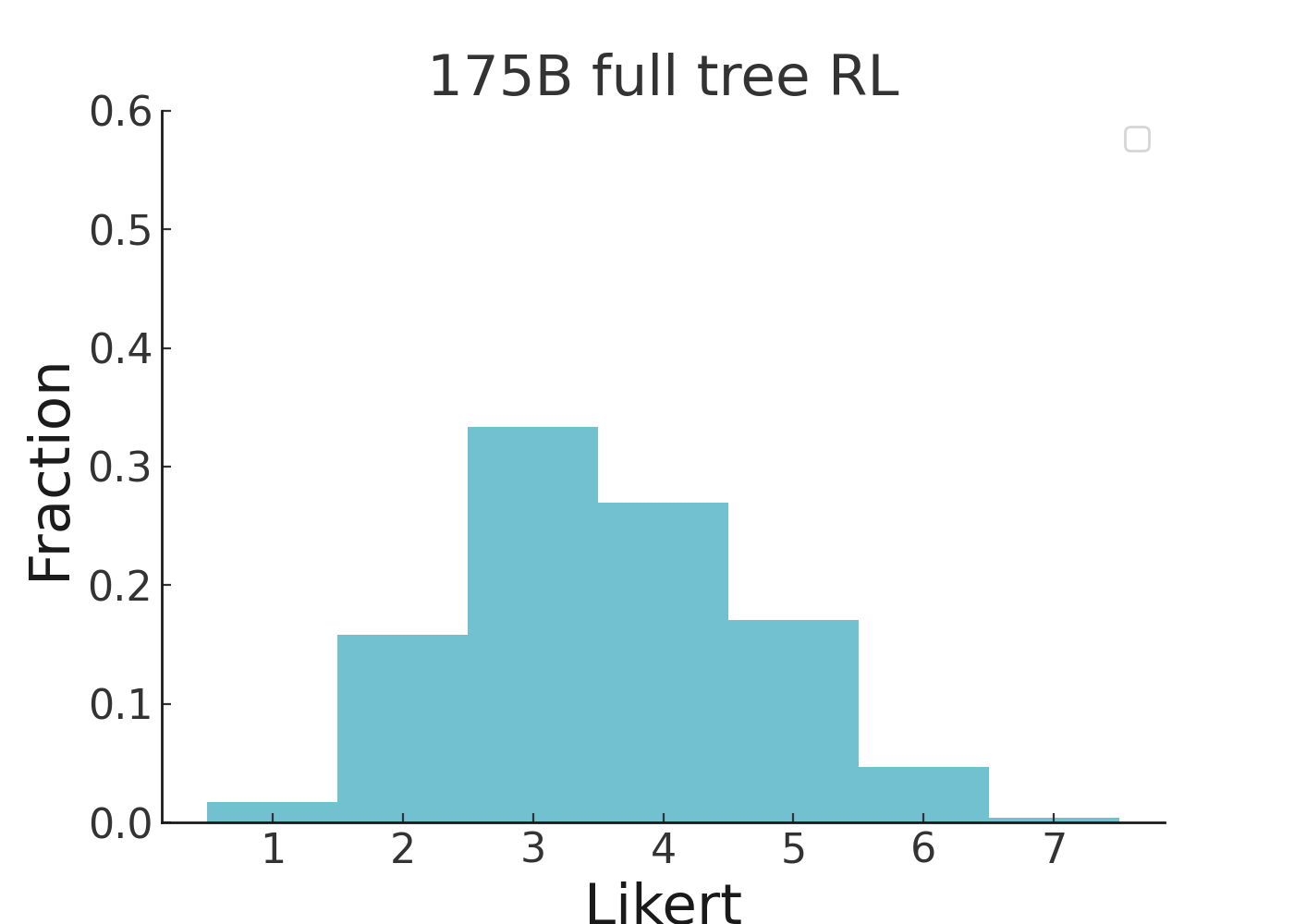}
    \newline
    \includegraphics[width=0.4\linewidth]{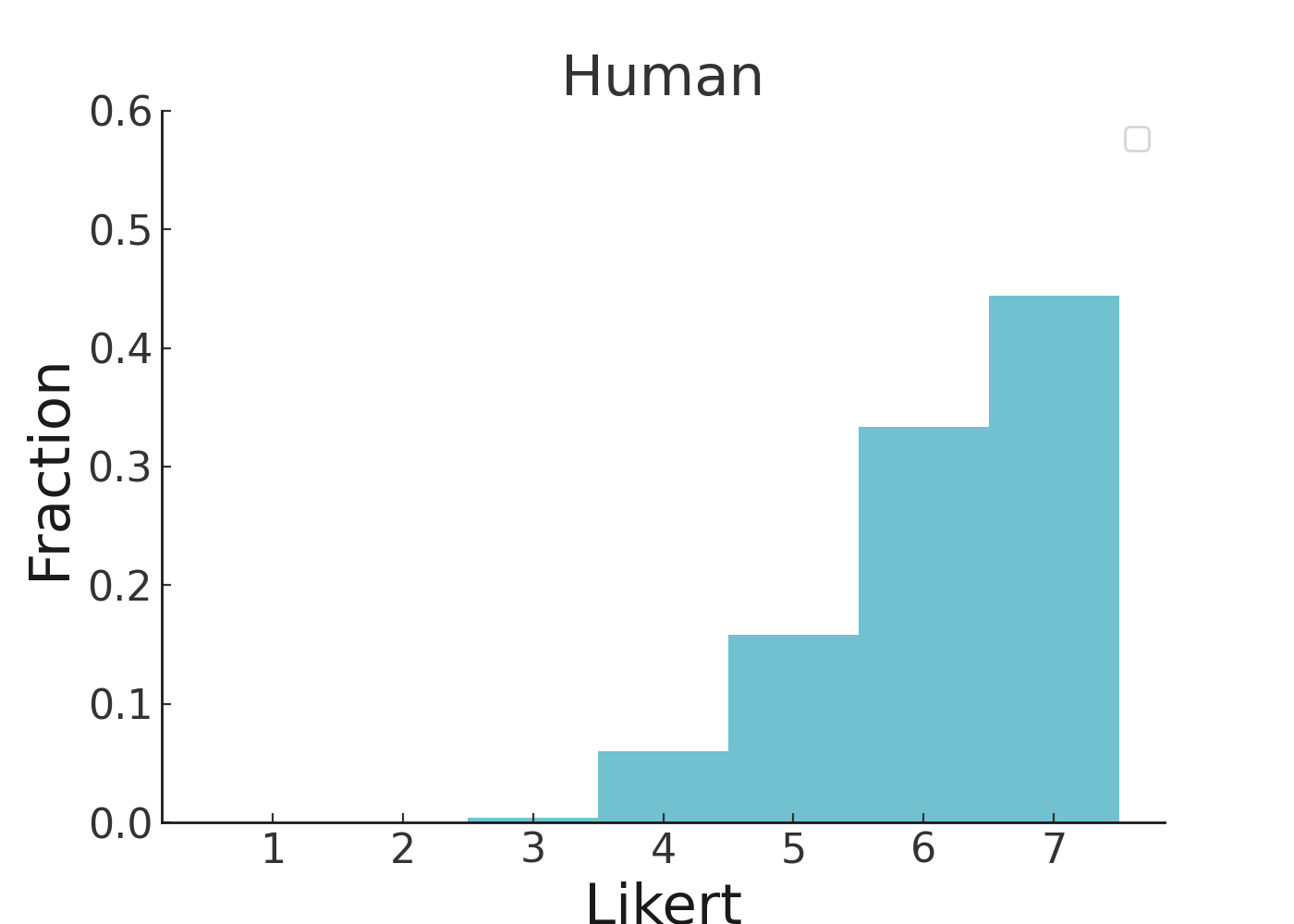}
    \includegraphics[width=0.4\linewidth]{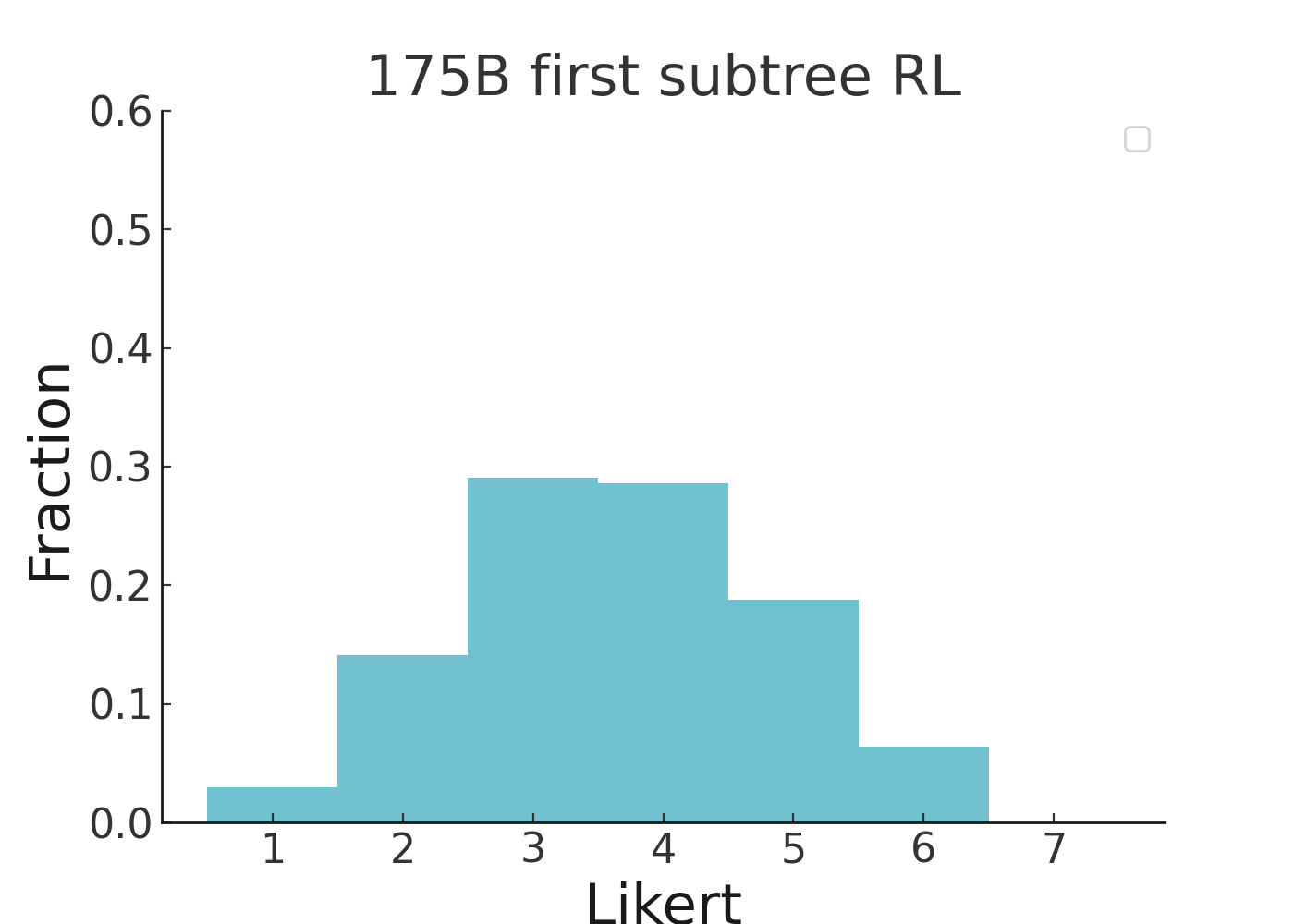}\hspace{8.5mm}

    \caption[]{\label{fig:fullbook_likert_hists}   Likert distribution for summaries from our supervised baselines (SL, same as BC), our best full tree RL models, our first subtree RL 175B model, and humans.
    }
\end{figure}

Our best models can generate realistic summaries of books unseen during training. Some of these summaries approach human-level quality: over 5\% of summaries from the best 175B model were given a score of 6 out of 7, and over 15\% were given a 5 out of 7, scores which were also sometimes assigned to human-written summaries (Figure \ref{fig:fullbook_likert_hists}).  However, on average our model summaries are still significantly worse than human-written summaries (Figure \ref{fig:fullbook_main}), 
See \href{https://openaipublic.blob.core.windows.net/recursive-book-summ/website/index.html#/goodreads}{our website}\footnote{\href{https://openaipublic.blob.core.windows.net/recursive-book-summ/website/index.html\#goodreads}{https://openaipublic.blob.core.windows.net/recursive-book-summ/website/index.html\#goodreads}} for our model summaries and ratings.

We find that training on the first subtree does comparably to training on the full tree (Figure \ref{fig:fullbook_human_time}).  Our models trained on just the first subtree generalize quite well to the full book summarization task.  However, we also found the full tree models disappointing; the final 175B full tree model we trained was noticeably worse than the previous one.\footnote{We had convincingly detected this prior to final evaluations via Likert scores for tree tasks, but included it for completeness.  The results in the remainder of the paper use the better (earlier) model, and we had committed to doing this before running final book evaluations.}  We discuss possible reasons for this in Appendix \ref{sec:fullrldifficulties}. We also find that our 175B RL policies significantly outperform our 175B BC baseline, though the improvement is smaller for the 6B models.

Likert scores for the full book summaries were significantly lower than Likert scores of any of the individual decomposed tasks.  This is unsurprising, since the errors accumulated at each depth are all reflected in the full book summary score.  See Appendix \ref{sec:true_task} for more discussion.

\subsection{BookSum results}
\label{sec:booksum}

We also evaluate our models on the recently proposed BookSum dataset for book-length summarization~\citep{kryscinski2021booksum}
We compare to the best extractive~(BertExt; \citealp{liu2019text}) and abstractive~(T5; \citealp{raffel2019exploring}) models, as well as an extractive oracle (which uses the reference summary to find the sentences in the source text that lead to the highest score). 
\citet{kryscinski2021booksum} evaluate book summaries using ROUGE~\citep{lin2004automatic}, BERTScore~\citep{zhang2019bertscore}, and SummaQA~\citep{scialom2019answers}.  SummaQA requires paragraph-aligned summaries, which we do not have, and so we report results on ROUGE and BERTScore.  Our depth 0 summaries are substantially shorter than the reference summaries, so we use the concatenation of depth 1 summaries.  

Our 175B models beat all non-oracle baselines on ROUGE by 3-4 points and approach the performance of an extractive oracle.  They also significantly outperform all baselines on BERTScore, including the extractive oracle. The 6B models are comparable to baselines on ROUGE while also significantly outperforming all baselines on BERTScore, including an 11B T5 model \citep{raffel2019exploring} fine-tuned on the BookSum dataset.

\begin{table}
\begin{center}

\begin{tabular}{ c  c c  c  c  c }
\toprule
& Abstractive & ROUGE-1 & ROUGE-2 & ROUGE-L & BERTScore \\
\hline
Extractive Oracle &  & \textbf{46.62} & 9.17 & \textbf{18.31} & 0.082 \\
\hline
BertExt &  & 36.71 & 6.16 & 13.40 & 0.028 \\
T5 zero-shot & \checkmark  & 35.43 & 5.62 & 12.02 & 0.011 \\
T5 fine-tuned & \checkmark & 39.46 & 7.69 & 13.77 & 0.060 \\
\hline
175b full tree RL & \checkmark & 41.51 & 10.46 & 16.88 & \textbf{0.1821} \\ 
175b first subtree RL & \checkmark & 43.19 & \textbf{10.63} & 17.10 & 0.1778 \\ 
6b full tree RL & \checkmark & 36.79 & 7.22 & 14.84 & 0.1246 \\
\bottomrule
\end{tabular}

\caption{\label{tab:booksum}Results on the test set of full book version of the BookSum dataset.  Baselines (top two sections) are from \cite{kryscinski2021booksum}. Our 175 RL models significantly outperform the non-oracle baselines.}
\end{center}
\end{table}

\cite{kryscinski2021booksum} report length being a confounder for BERTScore, with longer summaries having lower scores.  We also find a slight negative correlation between length and BERTScore, but controlling for it does not significantly affect our conclusions (see Appendix~\ref{sec:bertscore}).

Note that we cannot rule out overlap of the BookSum dataset with our pretraining dataset.  Nevertheless, from manual inspection of the trees, we believe that the summarization procedure largely reflects the structure of the book, rather than being a result of memorization from pretraining.

\begin{figure}
    \centering
    
    \subcaptionbox{}{\includegraphics[width=0.49\linewidth]{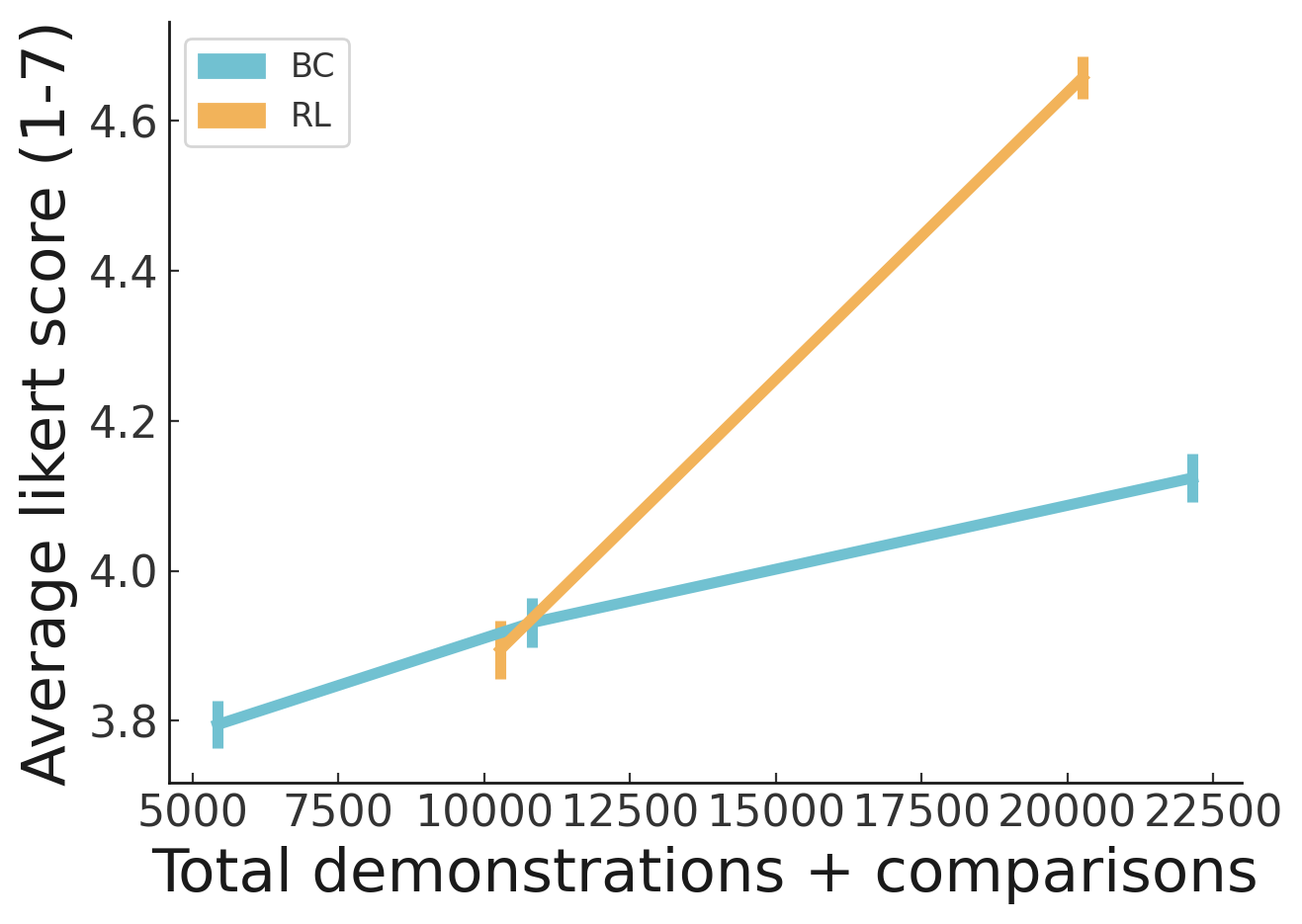}}
    \subcaptionbox{}{\includegraphics[width=0.49\linewidth]{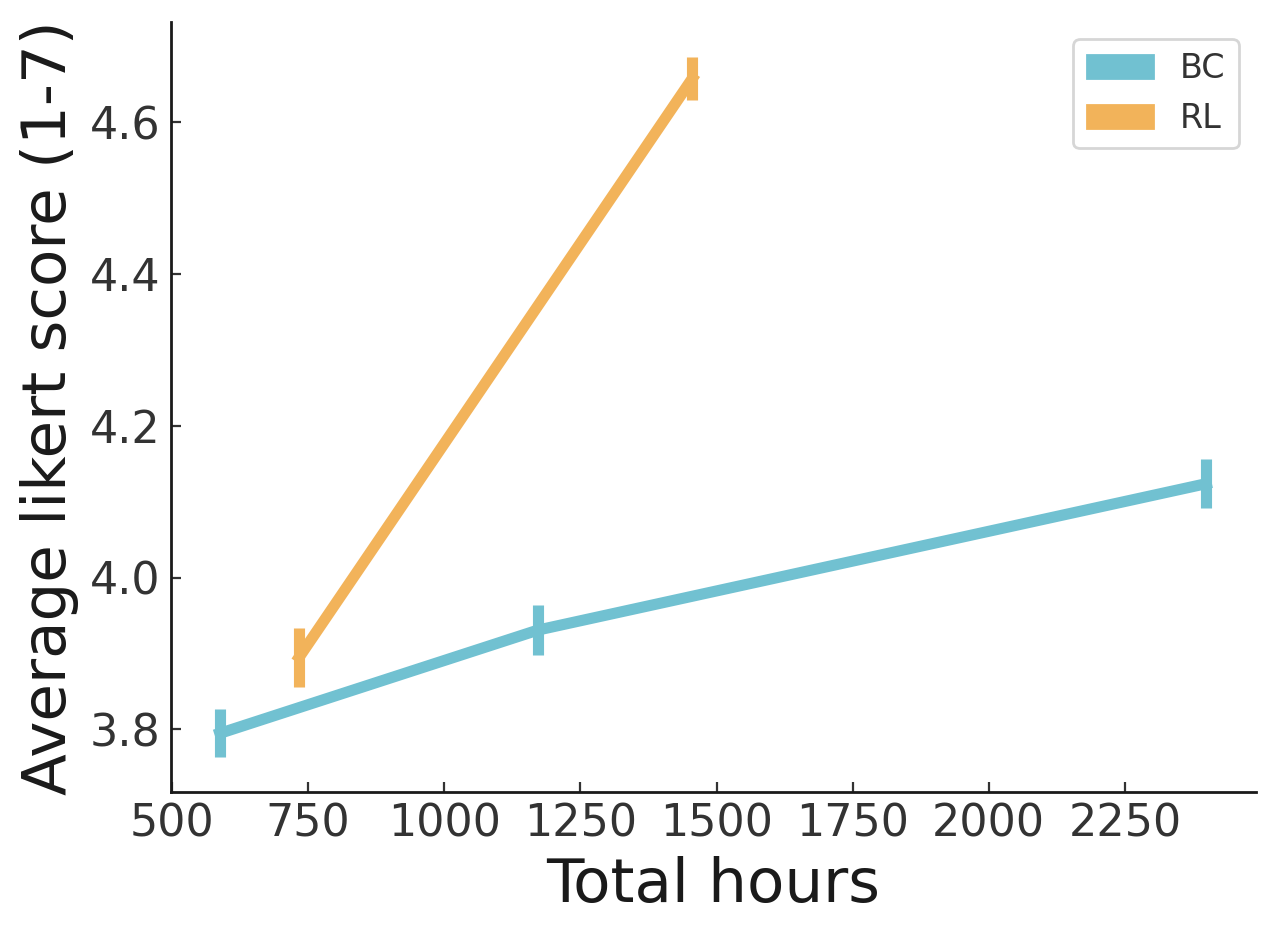}}
    
    \caption[]{\label{fig:human_time} (a) Performance on the first leaves, as a function of amount of human labels.  We see that there are diminishing returns to behavioral cloning, such that RL becomes substantially more efficient on the margin.  A policy trained with RL on 5K demonstrations + 5K comparisons is comparable to one trained with BC on 10K demonstrations.  However, one trained with RL on 10K demonstrations + 10K comparisons significantly outperforms one trained with BC on 20K demonstrations.  Standard error of the mean is estimated via bootstrapping at the label level.
    (b) Performance on the first leaves, as a function of amount of estimated human time.  Adjusting for human hours gives RL a greater advantage since comparisons are 3x faster to collect than demonstrations (see Appendix \ref{sec:humantime}).
    }
\end{figure}

\subsection{Human label efficiency of RL vs. BC}

In Section \ref{sec:fullbook_findings} we found that our RL models outperformed our BC models. However, our RL models were trained on significantly more data.
A significant open question is whether doing RL on summary comparisons is actually better than simple behavior cloning on an equal number of high-quality human demonstrations.
Previous results from \cite{stiennon2020learning} showed that doing RL greatly improved summary quality over their BC baseline, and even outperformed human-written summaries.  However, their reference summaries were scraped from Reddit TL;DRs, which are often not good summaries of the original text, and they do not compare to collecting a similar number of high-quality demonstrations.  

In this work, we use the same trained labelers to create demonstrations and comparisons, and directly compare RL to BC by plotting model performance versus the amount of human time required to produce each dataset. We study this on the first leaf summarization task rather than the full book summarization task to save human time.

We trained 3 versions of a 6B parameter BC baseline, with ¼, ½, and all the demonstrations.  Then, we trained RL policies starting from each of the ¼ and ½ BC policies,\footnote{We collected comparisons of the initial BC policies at temperature T=1, trained a reward model, and then ran a single round of RL with the initial BC policy at initialization.} with approximately the same number of comparisons as there were demonstrations. For these BC policies, we used temperature T=0.6, while for RL policies, we use T=0 (see Appendix \ref{sec:temperature} for justification).

We found that while RL on comparisons was about as effective as BC on demonstrations after 5k-10k demonstrations, comparisons were far more efficient on the margin after 10k-20k demonstrations (Figure \ref{fig:human_time}).  Furthermore, comparisons used to produce this figure were 3x as fast for us to collect as demonstrations (see Appendix \ref{sec:humantime}).

\begin{table}
\begin{center}

\begin{tabular}{ c  c  c  c  c }
\toprule
& ROUGE-L & BLEU-1 & BLEU-4 & METEOR \\
\hline
BiDAF \citep{kovcisky2018narrativeqa} & 6.2 & 5.7 & 0.3 & 3.7 \\
BM25 + BERT \citep{mou2020frustratingly} & 15.5 & 14.5 & 1.4 & 5.0 \\
RoBERTa \citep{zemlyanskiy2021readtwice} & 18.0 & 18.0 & 2.6 & 5.4 \\
ETC \citep{zemlyanskiy2021readtwice} & 18.8 & 17.2 & 2.7 & 5.4 \\
ReadTwice \citep{zemlyanskiy2021readtwice} & 23.3 & 21.1 & 4.0 & 7.0 \\
Retriever + Reader \citep{izacard2020distilling} & 32.0 & 35.3 & 7.5 & 11.1 \\
\hline
175b full tree, depth 1 & 21.03 & 21.82 & 3.87 & 10.52 \\
\hline
\textbf{\textit{6b}} full tree, depth 1 & 17.01 & 19.09 & 2.75 & 8.53 \\
175b \textbf{\textit{first subtree}}, depth 1 & 21.55 & 22.27 & 4.24 & 10.58 \\
175b full tree, \textbf{\textit{depth 0}} & 18.47 & 20.29 & 3.16 & 9.04 \\

\bottomrule

\end{tabular}

\caption{\label{tab:narrqa}Results on the test set for the full stories version of the NarrativeQA dataset. We ran a 3B UnifiedQA model against summaries from our best guess model (175b full tree with depth 1).  We also run ablations;  notably, our first subtree model outperformed the full tree model, consistent with results from Section \ref{sec:fullbookevals}}
\end{center}
\end{table}

\subsection{NarrativeQA: using book summaries for question answering}

Another way to evaluate summaries is to test whether they can be used to answer questions about the original text \citep{scialom2019answers,wang2020asking}.  

We applied our summarization model to the NarrativeQA question answering dataset \citep{kovcisky2018narrativeqa}, a dataset consisting of question/answer pairs about full book texts and movie transcripts.  The question/answer pairs come from Wikipedia summaries, matched by title to the full text.  In the full stories version of NarrativeQA, the model must use the original text.  

We test whether our summaries can be used as input (instead of the full book or movie text) to a question answering (QA) model.  For the QA model, we simply use a trained UnifiedQA model \citep{khashabi2020unifiedqa} in a zero-shot manner with temperature 0.  We can give it either the depth 0 summary, or a concatenation of the depth 1 summaries (the concatenation of depth 2 summaries can be quite long).  We found that depth 1 summaries work better.

As shown in Table \ref{tab:narrqa}, we achieve competitive results, despite our summarization model not being trained explicitly for question answering. However, we use far more parameters than \cite{izacard2020distilling}, the previous SOTA.   When using smaller UnifiedQA models for question answering, results are substantially worse, suggesting that the quality of the QA model is a primary bottleneck~(Figure~\ref{fig:narrativeqa}).  All our samples are available on \href{https://openaipublic.blob.core.windows.net/recursive-book-summ/website/index.html?data_id=175b\%2F0&dataset=narrativeqa#/narrativeqa}{our website}.

\section{Related work}

Our work is directly inspired by previous papers that lay the groundwork for applying human feedback to reinforcement learning~\citep{christiano2017deep}, especially to large-scale tasks. Our task decomposition approach can be thought of as a specific instantiation of iterated amplification~\citep{christiano2018supervising}, except we assume a fixed decomposition and start training from the leaf tasks, rather than using the entire tree. Similarly, our approach can be considered a form of recursive reward modeling~\citep{leike2018scalable} if we understand the purpose of model-generated lower-level summaries to be to help the human evaluate the model’s performance on higher-level summaries. Our contribution over these works is showing that this approach can be realistically applied to a difficult, large-scale task. We also build on the growing body of work that fine-tunes models with human feedback. This has been applied in many domains including summarization~\citep{bohm2019better,ziegler2019fine,stiennon2020learning}, dialogue~\citep{jaques2019way,yi2019towards,hancock2019learning}, translation~\citep{kreutzer2018can,bahdanau2016actor}, semantic parsing~\citep{lawrence2018improving}, story generation~\citep{zhou2020learning}, review generation~\citep{cho2018towards}, and evidence extraction~\citep{perez2019finding}, and agents in simulated environments~\citep{christiano2017deep,ibarz2018reward}. 

There has been relatively little work on summarizing novels and other long-form fiction writing. Early work~\citep{gorinski2015movie} used graph-based methods to summarize movie scripts. \citet{mihalcea2007explorations} introduced a dataset of book summaries scraped from CliffsNotes and tested an unsupervised extractive system based on MEAD \citep{radev2004mead} and Textrank \citep{mihalcea2004textrank}. More recently, \cite{ladhak2020exploring} propose a method for extractive summarization of chapters of novels. There has been work on generating partial summaries of fictional stories: \cite{zhang2019generating} investigate generating character descriptions written by the story author, and \citet{kazantseva2006approach} investigate extractive methods for generating information about the story setting and characters, but not the plot. Relatedly, \citet{bamman2013new} proposes an unsupervised method for aligning books with human-written summaries. There has also been some work on question answering using full books \citep{mou2020frustratingly,izacard2020distilling,zemlyanskiy2021readtwice}.  Concurrent with our work, \cite{kryscinski2021booksum} extended the datasets of \cite{mihalcea2007explorations} and evaluated neural baselines. 

While work on summarizing novels is sparse, there has been plenty of work on summarizing other kinds of long documents, such as scientific papers \citep{abu2011coherent,collins2017supervised,subramanian2019extractive,cohan2018discourse,xiao2019extractive,zhao2020seal,sotudeh2020generating}, and patents \citep{sharma2019bigpatent}, as well as multi-document summarization~\citep{liu2018generating,ma2020multi,gharebagh2020guir,chandrasekaran2020overview,liu2019hierarchical,gao2020supert}. Many of these techniques use a hierarchical approach to generating final summaries, either by having a hierarchical encoder 
\citep{cohan2018discourse,zhang2019hibert,liu2019hierarchical}, or by first running an extractive summarization model followed by an abstractive model~\citep{subramanian2019extractive,liu2018generating,zhao2020seal,gharebagh2020guir}. The latter can be seen as a form of task decomposition, where the leaf task is document-level extractive summarization and the parent task is abstractive summarization conditioned on the extracted summaries. 

The idea of decomposing hard tasks into multiple smaller sub-tasks has been used extensively in NLP. For example, \citet{fan2018hierarchical} generate fictional stories by first training models to generate a story prompt, and then training another model to generate the story conditioned on this prompt.  The idea of saving human time by using models trained at lower levels of the hierarchy to help humans label data for higher-level tasks has also been explored.  In \citet{fan2020generating}, models are used to search for evidence of facts, to help humans fact check faster and more accurately.

\section{Discussion}

Our main interest in this work is scaling human feedback to hard problems; we want to empower humans to give feedback to models on tasks that are very difficult to evaluate.
We expect this to be a critical part of the alignment problem because we need to make sure humans can communicate their values to AI systems as they take on more societally-relevant tasks~\citep{leike2018scalable}. If we develop techniques to optimize AI systems on what we actually care about, then we make optimization of convenient but misspecified proxy objectives obsolete.

In this paper, we showed that it is feasible to train models using human feedback on the difficult task of abstractive book summarization, by leveraging task decomposition and learning from human feedback. We also showed that doing RL on summary comparisons is more efficient than supervised learning on summary demonstrations, once the summarization policy has passed a quality threshold.
Though we used a fixed decomposition strategy that applies only to summarization, the general techniques could be applied to any task. In this sense we have made progress towards optimizing what we actually care about: good summarization performance as judged by humans. 

Something we do not address in this paper is training a single model to perform the entire top-level task, e.g.\ a single model that maps a book to a summary. This could be done via distillation as suggested in \cite{christiano2018supervising}, however in our case that would require training a single model with a very large context window, which introduces additional complexity. Furthermore, since the majority of our compute is at the leaf tasks, this would not save us much compute at test-time.

\subsection{Limitations}

\paragraph{Our model's book summaries lack coherence. } While our models successfully generate book-level summaries that contain much of the important information, they often read more as a list of events from the book, rather than a coherent summary that a human would write. In theory, this could be remedied with more rounds of RL at the top-level summarization task, however in practice we found RL at higher levels of the tree to be challenging (see below).

\paragraph{Task decomposition could be fundamentally limiting.} Task decomposition assumes that separate parts of the task can be completed independently. However, this may not be true for summarizing books. For example, it may be hard to catch cases where earlier details in the book are only later revealed to be important (e.g. in mystery books).  Our summarization models also sometimes generate inaccurate statements due to a lack of context; for example, there is a passage of Pride and Prejudice in which the main character gets asked for ``their hand''. In the broader context of the chapter, it is clear that the character is being asked for a dance. However, this is not clear from only the local context of the leaf task, and thus the model summarizes it as asking for ``her hand in marriage''. This is a general weakness of our training setup because we require each summary to be produced from only this local context, with a model that has not read the rest of the book.

Some of these issues may be alleviated by learning a decomposition procedure rather than using a fixed algorithm (see Appendix \ref{sec:true_task} for some discussion). However, this may not resolve all of the problems with decomposition.  Consider a case where important information is sprinkled lightly across many parts of the book, e.g. small details implying a buildup of love or resentment, where each detail is too minor to be included in a chapter summary despite being a prominent overall theme.  Determining the kinds of tasks that are amenable to decomposition remains an open problem.

\paragraph{Training on higher height tasks may be difficult.}

In general, policy errors at lower levels compound at each composition task, ultimately leading to large errors on the top-level task.  Auto-induced distributional shift (ADS, see Section \ref{sec:ads}) may also be making training significantly more difficult, and curriculum choice may matter a lot as a result.  Our curriculum and node sampling strategies were chosen in an ad hoc way. 

As shown in Section \ref{sec:fullbookevals}, training on the full tree of tasks did not lead to improved performance.  We discuss some possible reasons in Appendix \ref{sec:fullrldifficulties} but leave thorough investigations to future work.

\subsection{Open questions}

Though our approach produced plausible book summaries, the limitations above suggest some open questions for future research. First, are there better and more principled curricula?  Could one obtain improved performance by doing RL more on-policy, by generating the summary trees on the fly, or by training the reward model online as in \cite{ziegler2019fine}? Is it better to have longer or shorter episodes, encompassing more or less of the tree?  While having longer episodes means the policy has more in-distribution inputs at test time, it also means training on fewer trees for a given amount of compute and makes the reward model less on-distribution.

There are also many ways to improve the fundamental techniques for fine-tuning models using human feedback. For example, are there more efficient ways to collect data from humans instead of binary comparisons?
Could other methods for optimizing against human feedback, such as expert iteration~\citep{anthony2017thinking}, be more efficient? 

Finally, there are questions for how this procedure extends to other tasks. Is learning a task decomposition model, rather than using a fixed decomposition, feasible for hard real-world tasks? For what kinds of tasks is task decomposition fundamentally limiting? How else can we use ML models to assist humans in specifying their preferences for high-level tasks?  We hope to address some of these in future work.

\subsection{Broader impacts}

This work expands on the reward modeling technique proposed in \citet{ziegler2019fine} and \citet{stiennon2020learning}. Thus, the broader impacts are similar to the ones described in those papers. On the positive side, our research is motivated by the benefits of aligning ML systems with human intentions. We believe alignment techniques are an increasingly important tool to improve the safety of ML systems, particularly as these systems become more capable. Conversely, improved alignment could also enable malicious actors to more easily train models that cause harm, and could also lead to increased automation of some jobs, leading to job loss. See the broader impacts discussion of \cite{stiennon2020learning} for more discussion of these points. The difference in this paper compared to previous work on reward modeling is that we combine the technique with task decomposition, which allows us to use human feedback to train ML models to perform more difficult tasks. This amplifies both the potential benefits and the risks listed above.

One point we reiterate from \cite{stiennon2020learning} is to be careful when defining the `good' model behavior that labelers will reinforce. In other words, what or who should we align our models to? Deciding what makes a good summary is relatively straightforward, but defining good behavior becomes more difficult as we move beyond summarization to more complex tasks where humans might disagree on the correct model behavior. 

When solely considering the impacts of automatic book summarization, our models still make many mistakes while summarizing, and thus should not be deployed in a setting where high summarization accuracy is necessary. Our model summaries also seek to preserve the intent of the book, whose contents may be harmful or biased.

\subsection*{Acknowledgements}

We thank Wojciech Kryściński for discussion of book evaluation methods, and for help with BookSum;  Alec Radford for discussions about baselines and NarrativeQA;  Ben Mann, for help with our initial dataset;  Michael Petrov, Alethea Power, Chris Hesse, and the entire OpenAI Supercomputing team for help with infrastructure; and Alex Ray, Mark Chen, Tom Brown, Nick Ryder, and others for help with and work on pretrained models.

We also thank Jonathan Uesato, Ethan Perez, Sam Bowman, Wojciech Kryściński, and Diogo Moitinho de Almeida for detailed feedback and suggestions on the paper;  Pamela Mishkin for book suggestions and feedback on broader impacts;  Kelly Clancy for discovering the Pride and Prejudice example;  Natalie Summers for suggestions on books/scripts to use; Geoffrey Irving, Beth Barnes, William Saunders, and Dario Amodei for their support and thinking about our research agenda; Justin Wang for creating the graphics for the blog post; and  Jeff Clune for the idea to modify books to check prior knowledge.  

Last but not least, we'd like to thank all of our labelers, without whom this research would be impossible: Russell Bernandez, Gabriel Ricafrente, Laura Cowley-Martinson, Kelly Guerrero, Megan Niffenegger, Rachelle Froyalde, Ethan Myers, Stephen Ogunniyi, Jack Kausch, Jenny Fletcher, Charles Boone, Justin Dill, Celina Georgette T. Paglinawan, Bryce Vogel, Gabriel Perez, Cody St. Clair, Jelena Ostojic, Erol Can Akbaba, Maria Orzek, Alfred Lee, Ollie Horsfall, Eli Kapsack, Tasmai Dave, Cyra Mayell Denura, Sarah Mulligan, Emill Jayson Caypuno, Morris Stuttard, Ife Riamah, Sebastian Gonzalez, Vladan Djordjevic, Sarah Kirsten, Conor Agnew, William Brewer, Medeea Bunea, Joe Kwon, Chait Singh, Jennifer Brillo, Bashir Harrell, Leo Yung, Bekah Guess, Atresha Singh, and Jacob Bryan.

\newpage

\bibliographystyle{apalike}
\bibliography{references}

\newpage

\appendix

\addcontentsline{toc}{section}{Appendix} 
\part{Appendix} 
\parttoc 

\newpage

\section{Decomposition details and pseudocode}
\label{sec:decompdetails}

\subsection{Sectioning}
We generally aim for a text compression rate of 5-10x at each step, although the compression rate at top of the tree is typically lower, depending on the number of children of the root. 

We also generally aim to chunk text at white-space boundaries such as repeated newlines, chapter boundaries, etc., though we do not guarantee this and it is done heuristically.  

We filter out preamble and postamble using manually devised heuristics, though our labelers are instructed to output empty summaries upon such inputs if our heuristics do not catch everything.

Finally, the chunking code also consumes a random seed, allowing us to vary sectioning while chunking the above desiderata.

\subsection{Structure}

Inputs to leaf nodes are typically around 600 tokens.  Then, for height 1 tasks, we concatenate 10-13 summaries (each up to 128 tokens).  For higher height tasks, we target concatenating up to 8 summaries (each up to 192 tokens at height 2, or 384 tokens at higher heights), though it can be as low as 2 if there is not enough text, which is common at higher heights.

When applying our tree procedure, each book is split into about 200 leaf nodes on average, and about 20 height 1 nodes.  Trees typically reach height 3 (meaning there are additionally height 2 composition tasks, and a final composition task), but on rare occasions reach height 4 or greater.

\subsection{Using input model summaries as ground truth}
\label{sec:true_task}

For each task, we ask labelers to consider only the quality of the summary with respect to the direct input to the model, rather than the subset of the book representing the true summarization target.  Ideally, we would consider the ultimate task of the labeler or model to be to summarize or evaluate summaries of the full range of the book corresponding to the input in our decomposition.  The role of the existing best model would be as a "helper model" to aid in that task (by producing summaries of parts of the book), but the labeler/model would potentially still refer to the original text when needed.  Then the reward model at depth 0 would correspond to the "true" reward, rather than corresponding to only part of the trajectory.   

Had we defined the tasks this way, it may have helped address issues the error accumulation problem discussed in Section \ref{sec:fullbook_findings}.  When inputs were contradictory or confusing, labelers could consult the original source.  This would be particularly compelling if the model was also capable of question-answering.

Unfortunately, while we find this framing appealing, the pretrained models we had access to had limited context length.  Furthermore, this would have complicated our infrastructure and made the task for labelers somewhat more difficult.  Thus we start with the simpler version and leave such investigations to future work.

\subsection{General task decomposition pseudocode}
In this implementation of decomposition, the input at each step is simply a task which we wish to do, and a list of (subtask, response) pairs.  The subtasks are assumed to have come from a previous invocation of the function, and the subtask responses should help in answering the primary task.

\begin{python}
def do_task(task, subtask_pairs=[]):
     result = decompose_if_needed(task, subtask_pairs)
     if type(result) == Decompose:
         # recursively get the response to the subtask
         subresponse = do_task(result.subtask)
         return do_task(
             task, 
             subtask_pairs + [(result.subtask, subresponse)]
         ) 
     if type(result) == Respond:
         return answer_directly(task, subtask_pairs)
\end{python}

We have assumed existence of two functions:
\begin{enumerate}
    \item \verb|decompose_if_needed|, which returns either a \verb|Respond()| indicating the subtasks can be synthesized and answered by the model directly, or a \verb|Decompose(subtask)| if the model requires help to solve the task.  This subtask can be decomposed even further if necessary.
    \item \verb|answer_directly|, which returns an actual answer to the task, synthesizing the answers to subtasks
\end{enumerate}

In general, both \verb|decompose_if_needed| and \verb|answer_directly| could be learned and implemented by an ML model.  In the fixed decomposition case, \verb|decompose_if_needed| is implemented programmatically instead.

Note also that \verb|Decompose| only returns a single subtask, rather than a list of them.  This way, other child subtasks can depend on the result of the prior ones.

\subsection{Book decomposition pseudocode}
\label{sec:pseudocode_book}

A basic implementation of our tree decomposition for books described in Section \ref{sec:approach} might look like this:
\begin{python}
def decompose_if_needed(task, child_summaries):
    if len(task.text) < MAX_LENGTH:
        # just summarize actual book text
        assert not len(child_summaries)
        return Respond()
    # split text into parts of similar length
    chunks: List[str] = chunkify_text(task.text)
    # assume any existing N answers are for the first N chunks
    if len(child_summaries) == len(chunks):
        # we have all answers necessary, summarize concatenation
        return Respond()
    # We still need a summary for one of our children, recurse to it.
    # The outer loop will call the model to summarize this, 
    # and append to child_summaries
    return Decompose(Task(text=chunks[len(child_summaries)]))
    
def answer_directly(task, child_summaries):
    if not len(child_summaries):
        # actual book text
        to_summarize = task.text
    else:
        to_summarize = "\n\n".join(child_summaries)
    return model(to_summarize)
\end{python}

A version which correctly uses "previous context" is a bit more involved to implement. We keep an \verb|info| field which tracks a mapping from depth to all summaries written at that depth so far.  Note that the "previous context" summaries are from the same task depth (not necessarily the same task height).  For example, at height 0, if summarizing page 5-6, in addition to receiving the original text for pages 5-6, a model/human would also read the tail end of summaries for pages 1-4.  
\newline

\begin{python}
def decompose_if_needed(task, child_summaries):
    if len(task.text) < MAX_LENGTH:
        # just summarize actual book text
        assert not len(child_summaries)
        return Respond()
    # split text into parts of similar length
    chunks = chunkify_text(task.text)
    # assume any existing N answers are for the first N chunks
    if len(child_summaries) == len(chunks):
        # we have all answers necessary, summarize concatenation
        return Respond()
    # we still need a summary for one of our children, recurse to it
    new_info = add_context_info(task.info, child_summaries)
    return Decompose(Task(
        info=new_info,
        depth=task.depth+1,
        text=chunks[len(child_summaries)],
    ))
    
def answer_directly(task, child_summaries):
    if not len(child_summaries):
        # actual book text
        to_summarize = task.text
    else:
        to_summarize = "\n\n".join(child_summaries)
    return model(format_for_model(
        text=to_summarize,
        previous_context=get_context_for_depth(task.info, task.depth),
    ))
\end{python}

In words:
\begin{itemize}
\item{We are given a text to summarize, a depth $d$, and a mapping from depth to previous context for the text (which precedes the text we are summarizing)}
\item{If our text to summarize is small, we ask the model to produce a summary directly, conditioning on the previous context at our depth}
\item{If the text to summarize is long, we break it into N smaller chunks}
\item{We recursively ask for a summary of the first chunk, at depth $d+1$}.
\item{We append that first chunk summary to the previous context at depth $d+1$, and then recursively ask for a summary of the second chunk}.
\item{We repeat for all N chunks}.
\item{We finally concatenate the N chunk summaries into a final input, and summarize that, ensuring that the summary flows from the previous context at depth $d$}
\end{itemize}

\section{Labeler interaction details}
\label{sec:labelers}
\subsection{Selection and training}

We use a similar process to \cite{stiennon2020learning} for training labelers, and use many of the same labelers.  We pay labelers an hourly wage and have relatively extensive on-boarding materials.  All labelers are fluent in English and the majority are native speakers.

\subsection{Quality control}

We generally have a fairly involved quality control process, adopting techniques from \cite{stiennon2020learning}.   We often have a second labeler give detailed feedback on task completion, and give the first labeler a chance to respond to that feedback.  When doing composition tasks with human-written inputs, we also give a chance for labelers to give feedback on those inputs.

We also communicate frequently with our labelers via Slack, giving them a chance to give us feedback and vice versa.

\subsection{Task interface}

\begin{figure}
    \centering
    \includegraphics[width=0.8\linewidth]{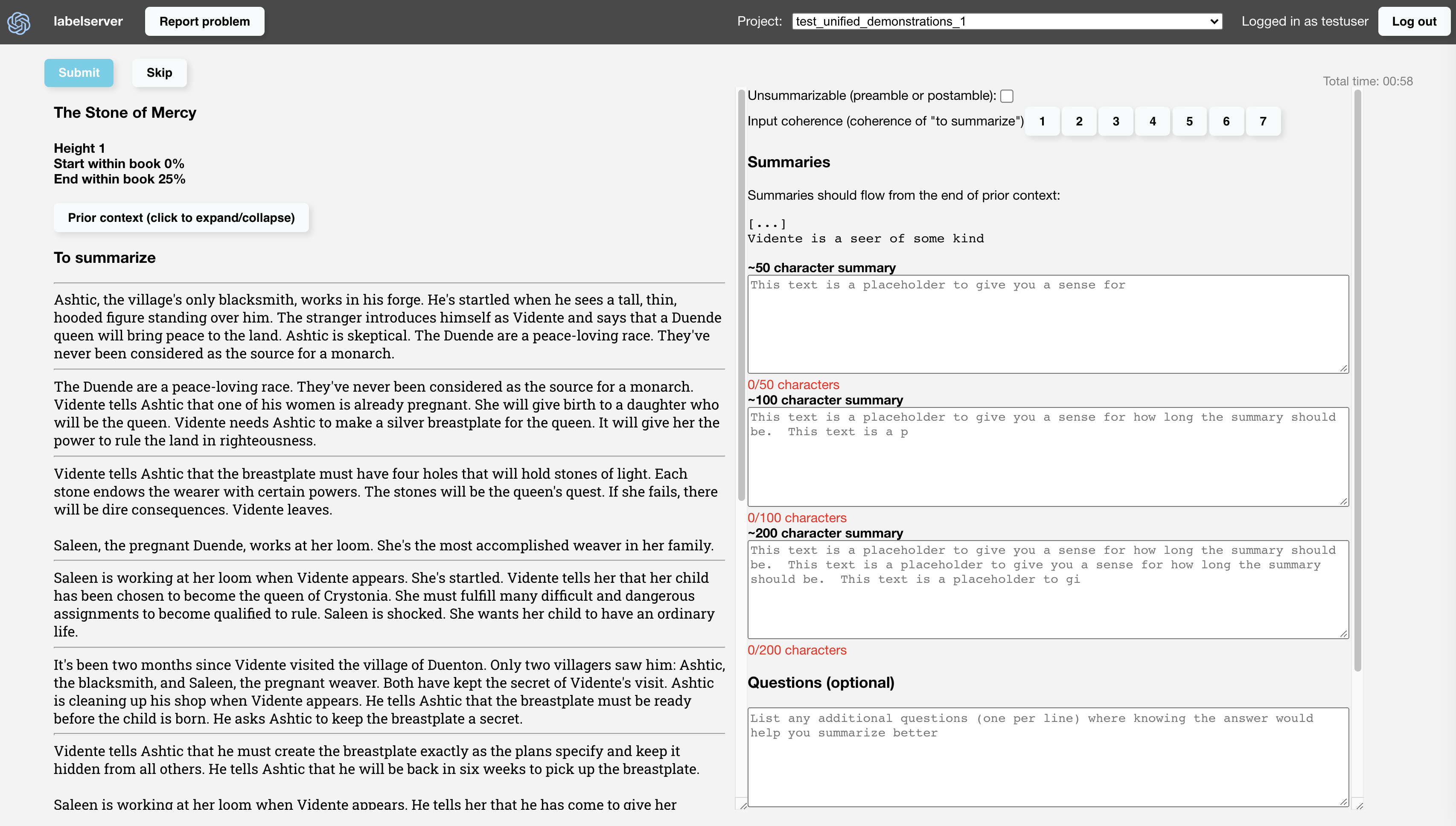}

    \caption[]{\label{fig:labelserver} Renderer for producing summary demonstrations.
    }
\end{figure}

We use a website and task-allocation library developed specifically for giving tasks to labelers.  We use different customized "renderers" for different tasks (demonstrations, comparisons, final evaluations, etc).  See Figure \ref{fig:labelserver} for an example of a demonstrations renderer.

\section{Labeling task details}

\subsection{Guidelines}
\label{sec:labeler_guidelines}

The following guidelines were given to labelers for evaluating summary quality, and applied to both demonstrations and comparisons.

We have three primary criteria

\begin{enumerate}
\item \textbf{Coverage}: All information in the summary should be important, and there should be no other more important information omitted from the summary.  So gratuitously including small details is generally penalized, and omitting important details is also penalized.  
\item \textbf{Accuracy}: All information in the summary should faithfully reflect the original passage.
\item \textbf{Coherence}: Ignoring the passage, the summary should not be confusing, ambiguous, or logically incoherent.
\end{enumerate}

We also have a fourth criteria which is primarily applicable at higher height.  Labelers were to use their own judgment on how important it was
\begin{enumerate}
  \setcounter{enumi}{3}
    \item \textbf{Abstraction}: When possible, writing should describe larger arcs and themes rather than just listing a series of events that happened.
\end{enumerate}

In addition, we also have the following guidelines
\begin{itemize}
\item The summary should flow from the end of the previous context
\item When using pronouns, resolutions should be clear for a naive reader
\item Present tense should be preferred
\item Reader uncertainty should be indicated in square brackets, e.g. [maybe]
\item Line breaks should be used to indicate a change of scene
\item Output should be empty if the content is preamble/postamble (publishing details, etc.)
\end{itemize}

\subsubsection{Length}
\label{sec:length}
 
Comparing summaries of different lengths can be very difficult, and result in e.g. systematic preferences for longer summaries, if labelers value summaries being informative rather than concise.  Length was found to be a significant confounder of quality in \cite{stiennon2020learning}, who report length-controlled results.  

Consistent with our coverage criterion, we ask for the best summary “overall”, controlling for length -- a summary is evaluated for the particular length it was written at.  For example, if summary A was 100 tokens and summary B was 200 tokens, we asked labelers to imagine that summary A had a 100 token “budget”, summary B had a 200 token “budget”, and to report which summary did a better job of using its budget.  Overall, in our work, we find length has an insignificant effect on summary quality.  This avoids the need to control for length.  

Nevertheless, we set limits on length.  Our allowed range of lengths increase as we summarize more of the book.  We institute hard limits of 128 tokens for the height 0 (leaf level) tasks, 192 tokens for height 1, and 384\footnote{We increased the limit mid-project from 192, and typical lengths are still much closer to 192.} for all other heights.  In practice, we do not frequently hit these length limits - when they are exceeded, we truncate the summaries before they are shown to humans (and before shown in this paper).

\subsection{Differences between human and model tasks}

In principle, our models and humans should be performing the exact same task.  In practice, they differ very slightly, though we expect none of these differences affect results or conclusions

\subsubsection{Demonstration lengths}
For demonstrations, although we ask for best “overall” taking length into account, humans can just as easily write good summaries at different lengths.  Thus we gave our labelers a range of different suggested length targets within the acceptable range, with 20\% headroom in either direction.  This ensured our models tried outputting summaries at different lengths.  The suggested lengths are typically chosen between half the length limit and the limit, roughly between 100 and 200 BPE tokens.  

\subsubsection{First leaves “contamination”}
\label{sec:contamination}

When collecting data (demonstrations and comparisons) on the first leaves, we typically have labelers do all the tasks consecutively at once, thus saving a bit of time by virtue of already having paged in the previous context -- though this did cause labelers to see more context than a model doing the same task saw.  

When doing the “contaminated” comparisons, labelers typically saw the same previous context for the summaries being compared.  However, for some period, our reward model was seeing summaries with different previous contexts (for the same data collected).  

\subsubsection{Comparison amortization}

Much of the expense of collecting a comparison is in reading the input text.  We can speed up comparison collection by asking labelers to compare multiple pairs of summaries for each input text (at the cost of higher correlations in the collected data).  Furthermore, the pairs of summaries can have overlap.  In practice, we use up to 3 pairs of comparisons between 3 summaries.  Though we could use a similar trick for demonstrations, we tried it briefly and abandoned it, as we were afraid the demonstrations for the same text would be too similar when written in quick succession.

\subsubsection{Additional data collection}

For valuation and diagnostic purposes, we also collect the following data, at various points in time:
\begin{itemize}
    \item When doing comparisons of summaries, we collect 1-7 Likert ratings for the primary criteria mentioned in Appendix \ref{sec:labeler_guidelines}.  We also always collect an overall Likert rating.  Ratings reflect absolute quality rather than relative quality (to another summary).
    \item We also ask for ratings of coherence of the \textit{input texts} for composition tasks
\end{itemize}

At various points in time, we also collected other datasets, including but not limited to:
\begin{itemize}
    \item Annotations of spans in the summary which were inaccurate, incoherent, or exhibit poor coverage
    \item Questions about the texts being presented
    \item Various free-form notes on the task
\end{itemize}

Overall, none of these data affected the primary task (of demonstration/comparison) in any way, and were simply supplementary data intended for future experimentation.

\section{Additional training details and hyperparameters}
\label{sec:finetuning_hyperparameters}

\subsection{Fine-tuning details}

Our hyperparameter choices follow those of \cite{stiennon2020learning}.  BC models and reward models are trained for 1 epoch, with cosine decay.  Learning rates are chosen by a separate sweep for each model size, and we use a cosine decay schedule.  We use the Adam optimizer.  

Like \cite{stiennon2020learning}, for reward models, we add an additional head on top of the final layer, initialized randomly.  We often run multiple seeds and choose the best reward model based on validation loss/accuracy.  We normalize the reward model to be zero-centered around human demonstrations prior to using it for RL.  This makes it slightly easier to compare rewards across runs, and likely affects the optimization in a beneficial way (if at all).  We also initialized the value function to the reward model weights, which we found helps learning.

For reinforcement learning, we primarily tune KL coefficient and learning rate.  KL coefficient is generally chosen in an ad-hoc way to target a KL range we deemed reasonable - we used 0.02 for most runs, but also experimented with 0.01 and 0.03 earlier in the project.  Learning rates are chosen using sweeps for each model size (very roughly chosen, for 175B).  We use linear learning rate decay and run for up to 200,000 episodes (for most of the project, we used 150,000 episodes).

\begin{figure}
    \centering
    \includegraphics[width=0.6\linewidth]{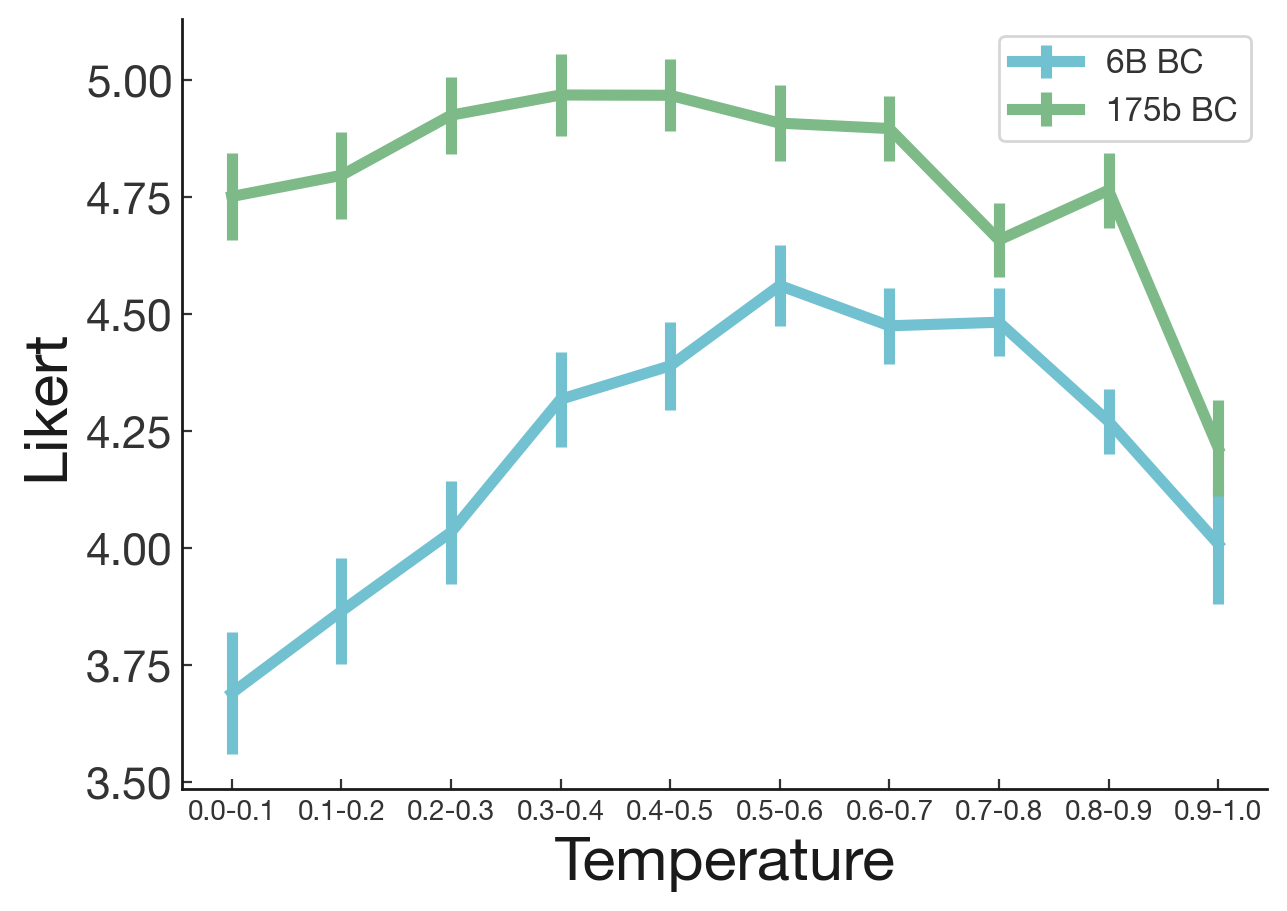}
    \caption{\label{fig:temp_sweep}Likert values at different temperatures, for a 6B and 175B supervised model on leaf tasks.  Standard error estimated via bootstrapping at the label level. \vspace{-5mm}}
\end{figure}

\subsection{Temperature}
\label{sec:temperature}

To ensure that we compared against a fair baseline, we swept temperatures and had labelers evaluate quality of various BC models on the leaf tasks.  In Figure \ref{fig:temp_sweep}, we find that the 6B supervised model is best at around T=0.6, while the 175B supervised model is best around T=0.3.  

Higher level tasks followed similar overall pattern (although we have noisier estimates).  We later found in final evaluations that better temperatures for individual tasks was predictive of performance on the full book summarization tasks as well.

\subsection{Input format}
The input format concatenates the following, in order:  previous context summaries separated from each other by "\verb|\n----\n|", the separator "\verb|\n====\n|", the text to summarize, and finally the phrase "\verb|TL;DR:|".  The model then generates the summary after that.  The previous context summaries are truncated (from the beginning) to fit within the 2048 token context window while leaving room for a summary of maximal length.

\section{Human timing}
\label{sec:humantime}

\subsection{First leaves}
We collected detailed timing information which let us know how long the primary tasks took.
  
We found empirically that comparisons are about twice as fast as demonstrations, ignoring read time.  Including read time, they were about 40\% faster.  For leaf tasks, where the distribution is not policy-dependent, we estimate 2.5 minutes reading, 4 minutes per written demonstration, and 1.5 minutes per comparison.  

Since for both (especially comparisons), reading the passage is a non-trivial part of the cost, amortizing the read time across many demonstrations or comparisons can help increase rate of data collection. We briefly tried collecting demonstrations of different lengths;  however, we found the demonstrations to generally be quite similar and stopped collecting such data early on.  For comparisons, we typically collect 3 at a time, thus amortizing the read time down to around 0.8 minutes for leaf tasks. This makes comparisons nearly 3x faster than demonstrations (2.3 minutes vs. 6.5) minutes.  Empirically, we find it over 3x faster (1.8 minutes).  This may be because we typically compare all pairwise combinations between 3 samples, thus yielding only $\log_2(6) = 2.58$ bits of information rather than $3$ bits, but also saving on time processing each summary.  Similar results hold across all heights.  Demonstrations generally took between 10 to 15 minutes total, while a set of 3 comparisons also took between 10 to 15 minutes.

Our results in Figure~\ref{fig:human_time}a on the first subtree uses these practices.  The results hold despite comparisons being 3x faster to collect and each yielding far less information (less than 1 bit per comparison, versus potentially thousands per demonstration).  When plotting with estimated human time, the advantage of RL is more apparent, see Figure~\ref{fig:human_time}b.

\subsection{End-to-end baseline estimates}
\label{sec:humantime_estimates}

It took over 12 hours on average for a labeler to read a full book, and additionally over 1 hour to write the summary.  This is over 50 times longer than it takes labelers to do a single decomposed summarization task.  Thus using the same amount of human time as Figure \ref{fig:fullbook_human_time} (enough for 100K total demonstrations and comparisons), we would have had summaries for at most 2K distinct books.  

While existing datasets of book summary datasets can be scraped from the Internet (e.g. from study guides such as Sparknotes), they typically have only hundreds of well-known books.  For example, \cite{bamman2013new} has 439 (book, summary) pairs.

Another consideration is that reading time can be amortized greatly by having contractors write multiple summaries per book.  In practice, we found it difficult to have contractors write multiple distinct summaries.  Nevertheless, this could plausibly save a substantial amount of time if executed well.  

Furthermore, learning the book summarization task end-to-end would likely be much more difficult than the decomposed tasks, as the model would need to learn attributions across an extremely long context.  Overall, we believe an end-to-end baseline would likely have been infeasible with naive methods, but leave it to future work to try.

\section{Mistakes and miscellaneous learnings}

\subsection{Mistakes}
\begin{itemize}
\item Given that we were doing a recursive strategy, we should’ve made the base case smaller.  Reward modeling did not work as well on the leaf level task as it had on the TL;DR task from \citet{stiennon2020learning}. With shorter input texts and summaries, we may have seen signs of life much sooner.
\item The “contamination” set up (see Appendix \ref{sec:contamination}) complicated our infrastructure, and resulted in task mismatch between the human and model.  We likely should have assumed the more general set up immediately.
\end{itemize}

\subsection{Miscellaneous Learnings}
\begin{itemize}
\item We tried initializing reward models from the previous one and fine-tuning on only the data collected since.  We could not tell whether this was better or worse, though it saved on compute.
\item Similarly, we considered initializing RL models from the previous one (and also using the previous RL model for the KL penalty).  However, RL seems to lose entropy in suboptimal ways: at some point, our model really favored summaries that started with "[X] reflects".  For this reason, we always use the most recent supervised policy, rather than the best RL policy, for the RL initialization and KL penalty.  However, further investigation is needed.
\item We collected structured feedback of when the models made coverage/coherence/accuracy mistakes, with highlights of spans where errors occur.  Training on this data as a supervised task did not help as initialization for reward models.  However, this was very exploratory and we remain very excited about future work in this direction.
\item Postamble filtering didn’t seem necessary, even though the model was barely trained on postambles (whereas comparatively a lot of training data contained preambles)
\item Training a reward model to directly predict Likert scores using a least squares lost resulted in similar accuracy to our binary comparison based models.  
\end{itemize}

\section{Difficulty and mysteries of full tree training}
\label{sec:fullrldifficulties}

As shown in Section \ref{sec:fullbookevals}, training on the full tree of tasks did not lead to improved performance. We give some possible reasons for this.  

\begin{enumerate}
    \item \textbf{Lack of hyperparameter tuning}:  We did not tune the 175B models much due to compute costs.  
    \item \textbf{Poor input distribution and noisy comparisons for higher level tasks}:  The quality of the input summaries given to the model (and thus to human evaluators when evaluating this model) degrades as one moves up the tree. The quality of input summaries is important for labeling accuracy: we found that inter-labeler agreement went down when labelers judged the input summaries as less coherent. Thus, the training signal degrades if we move to training on higher level tasks too early, before the summarization models have passable summaries.
    \item \textbf{Poor node sampling during RL}:
    Our episode sampling strategy described in Section \ref{sec:rl_variants} may have been suboptimal.  Rather than the vast majority of tasks being height 0 tasks, only about one third are.  This is in contrast with evaluation time, where height 0 are both most numerous and potentially most important.  Empirically, we found that the best full tree 175B RL model did sacrifice performance on lower heights in order to do better at the higher height task, relative to the best first subtree model (which had similar full book performance overall).  However, the later full tree 175B RL model, shown as the unfortunate dip found in Figure \ref{fig:fullbook_human_time}, had worse Likert scores at all heights.  This makes the explanation somewhat unlikely, although it is possible that it is actually better at higher heights and the shift in lower height summaries makes it appear worse.
    
\end{enumerate}

Most of the below reasons do not explain why training on more full tree data decreased performance for our 175B model.  We do not have good hypotheses for why, but we also cannot rule out a bug in the training code, or randomness across RL runs. Our initial guess was that the behavioral cloned model or the reward model performance had degraded -- however, they did not regress significantly on lower height tasks on loss and accuracy metrics, compared to corresponding models trained only on first subtree data. While this does not rule out a reward model which is generalizing worse in some way during RL, it leads us to believe the issues were primarily elsewhere in the RL.

\section{NarrativeQA: additional findings}
\label{sec:narrativeqa_otherfindings}

\subsection{Ablations}

\begin{figure}
    \centering
    \includegraphics[width=0.49\linewidth]{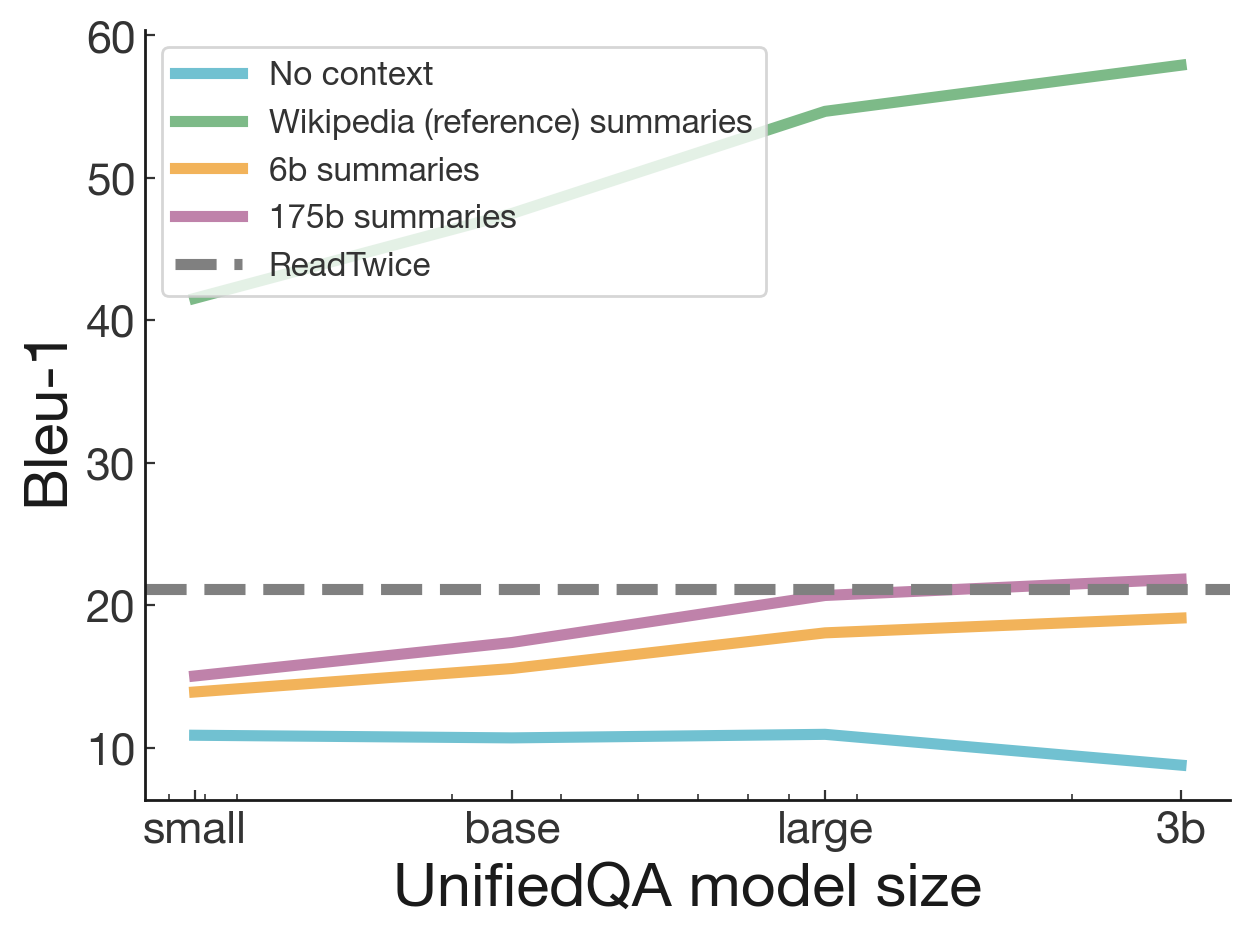}
    \includegraphics[width=0.49\linewidth]{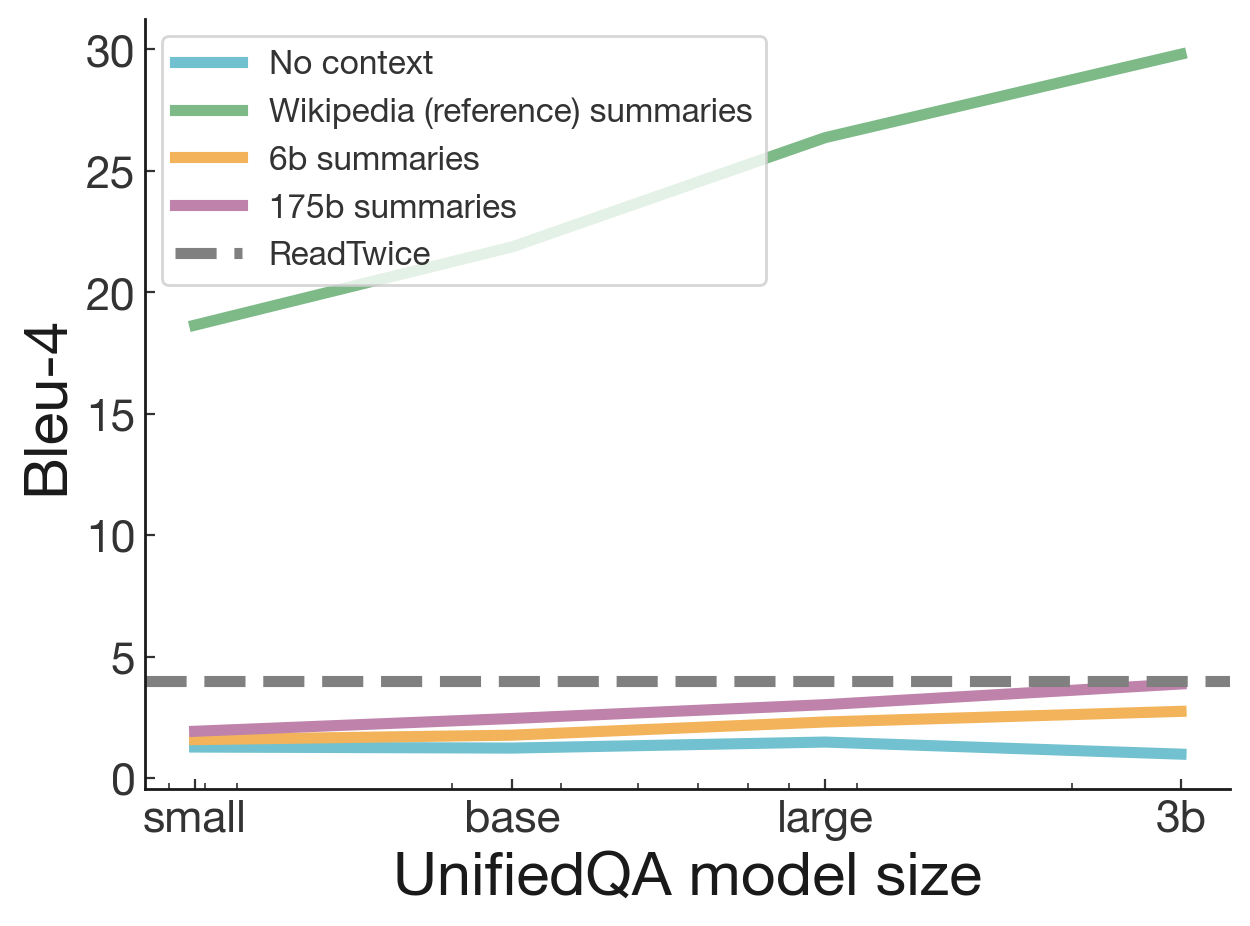}
    \newline
    \includegraphics[width=0.49\linewidth]{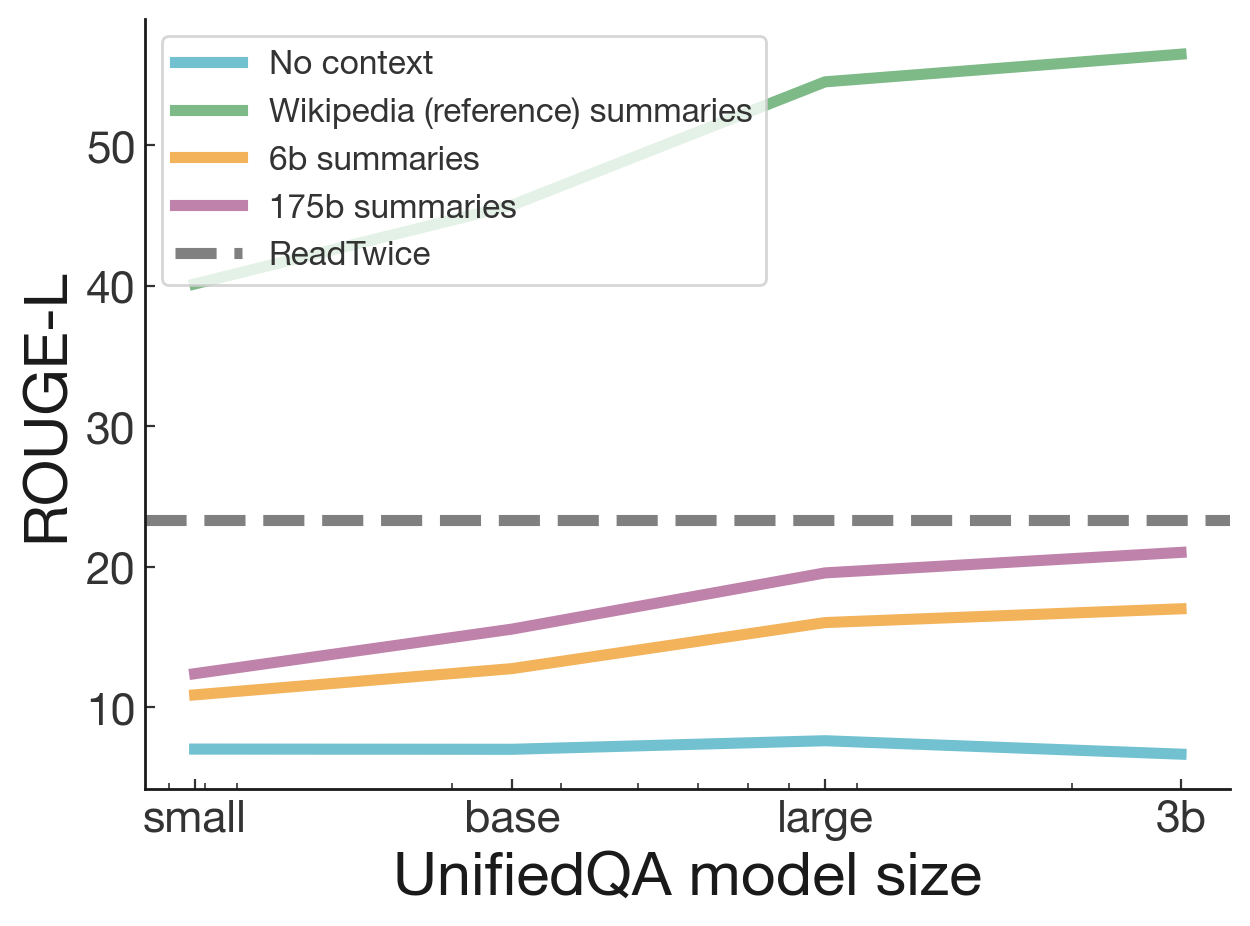}
    \includegraphics[width=0.49\linewidth]{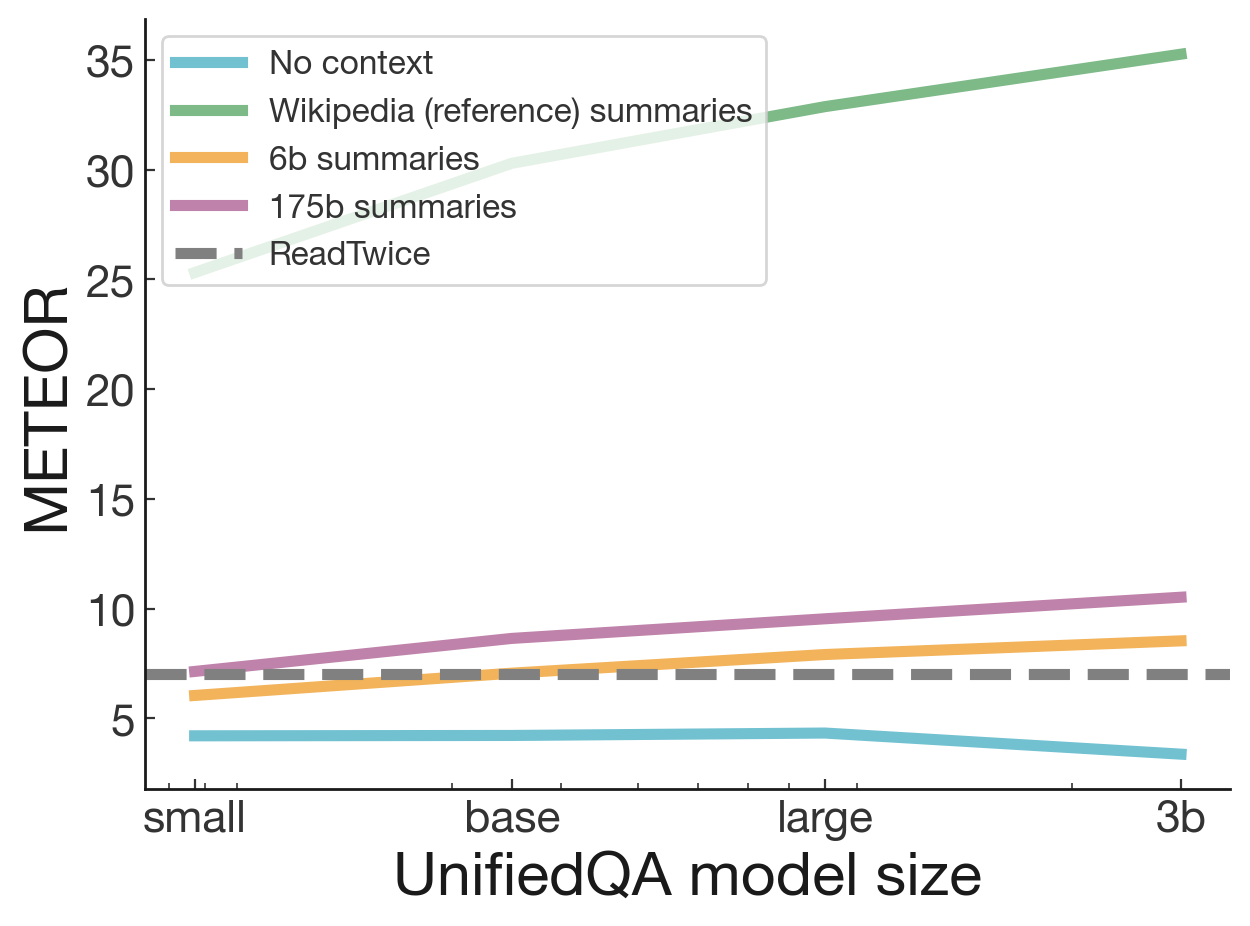}

    \caption[]{\label{fig:narrativeqa}   Various ablations on NarrativeQA, using depth 1 summaries from models trained on full trees.  Ablations of QA model size: we still see strong returns on QA model ability with all summaries (except empty ones).  Ablations of summary context: empty contexts provide a lower bound on performance, and NarrativeQA's reference Wikipedia datasets provide an upper bound. 
    }
\end{figure}

In Figure \ref{fig:narrativeqa} we show ablations on NarrativeQA for both the UnifiedQA model size and the summaries used as input to the UnifiedQA model.

\subsection{GPT-3 memorization}

For the QA model, we also attempted using a pretrained GPT-3 in a few-shot manner, similar to \cite{brown2020language}.  However, unlike the UnifiedQA models, pretrained GPT-3 surprisingly achieved extremely strong performance without any summaries.  In fact, the 175B parameter model had state of the art results according to all metrics except ROUGE-L (which was extremely close).  

\subsection{Zero-shot recursive question answering}

We also attempted a method of prompt engineering to cause our summarization model to act as a recursive question-answering model.  To do this, we run our tree procedure, but augmenting each step with the question.  Specifically, we add an additional prompt between the passage and response: "Answer the following question based on the above passage, or reply with a summary of relevant information if no answer is found: \{question\}".  The procedure can be viewed as a type of summarization with respect to a question.  Unfortunately, this is quite expensive, since we need to re-run the entire tree for each question.  

In a small sample of 100 such depth 0 trees produced this way, the authors found this gave even better answers, although the "answers" tended to still include extraneous summarization-like information.  The authors found 29 of 100 questions were correctly answerable (clearly agreed with at least one of the gold labels), and a further 8 were either partially correctly answerable or correctly inferrable.  On the other hand, for the trees without the question augmentation, we deemed only 10 of 100 correctly answerable, and 12 partially correctly answerable or correctly inferrable.  During this process, we also found that a substantial percentage of the NarrativeQA dataset appeared to have incorrect texts, where the questions do not appear to be about the correct book.

\subsection{Comparison to prior work}
\label{sec:narrqa-prior}

The NarrativeQA results highlight that our model summaries contain enough useful and accurate information to answer questions about the original book. 
While previous methods are far more parameter efficient (\cite{izacard2020distilling} had 2 orders of magnitude less parameters and ReadTwice \citep{zemlyanskiy2021readtwice} had nearly 3 orders of magnitude fewer parameters), there are some advantages of using an approach like ours:
\begin{enumerate}
\item First, our technique is quite general, and answers questions fully abstractively, rather than via token extraction.  For example, we observed the model inferring that a country of interest was England, despite it having no explicit mention in the summary besides the mention of London.  
\item Second, when answering 30 questions per passage, we require only one forward pass over the full book rather than 30, with the remaining passes being over a much smaller text.  (On the other hand, we cannot answer questions that are not answered by the summary.)
\item Lastly, and most importantly, we retain the benefits of decomposition.  Our model’s answers can often be easily traced back to the source in the book, and by leveraging the tree structure, we can often tell where mistakes led to wrong answers. Our model’s summaries can help a human perform question answering quickly -- see Appendix \ref{sec:narrativeqa_otherfindings} -- whereas the approach of \cite{zemlyanskiy2021readtwice} produces hard-to-interpret latents.  
\end{enumerate}

\section{BookSum: BertSCORE length control}
\label{sec:bertscore}

\begin{figure}
    \centering
    \includegraphics[width=0.7\linewidth]{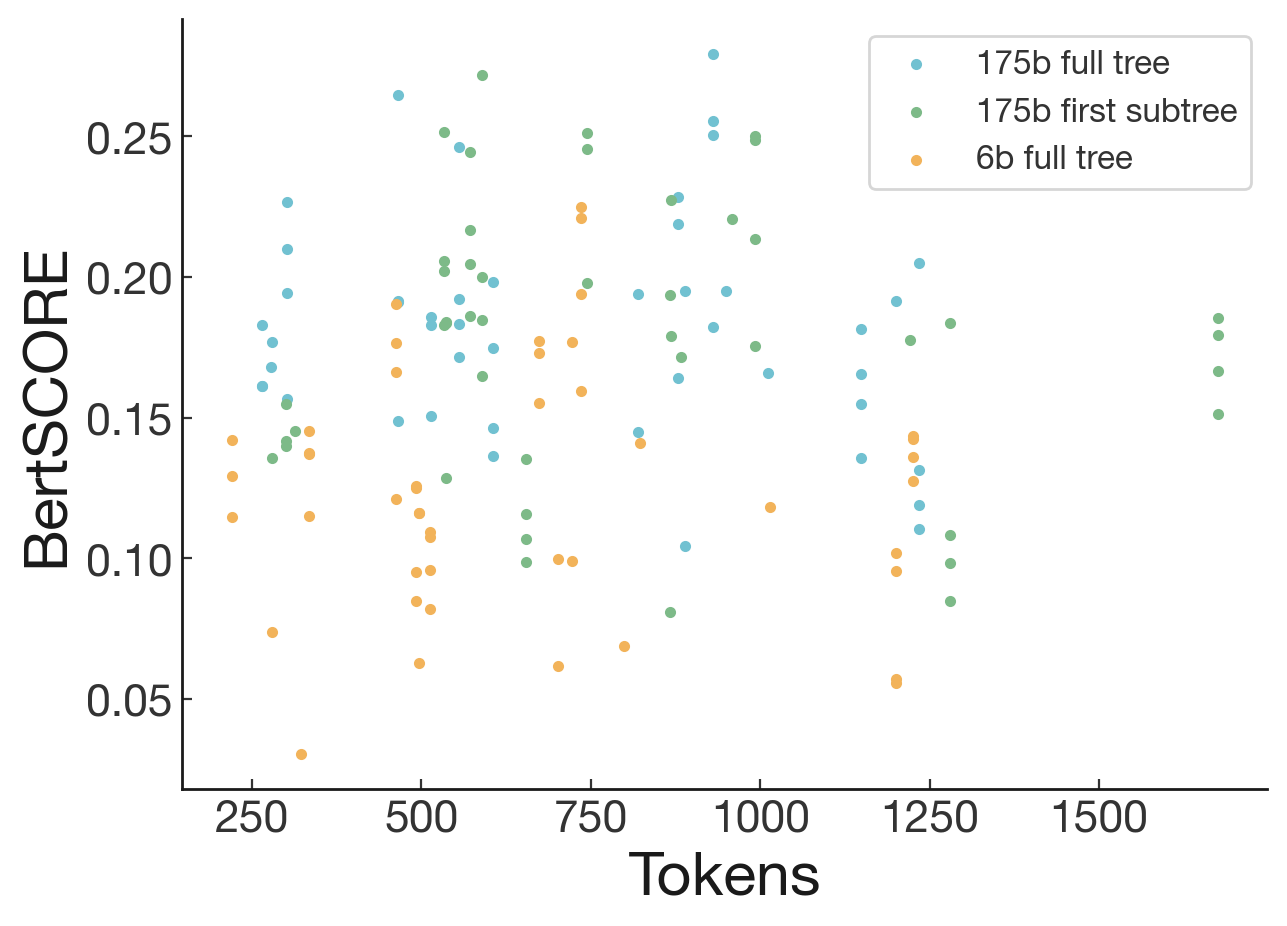}
    \caption{\label{fig:bertscore_length}Scatter plot of BERTScore vs length, for all 46 books in the BookSum test set.  
\vspace{-5mm}}
\end{figure}

\cite{kryscinski2021booksum} report length being a confounder for BERTScore, with longer summaries having lower scores.  We also find a slight negative correlation between length and BERTScore.  However, using a simple linear regression to control for length does not significantly change our scores.  See Table \ref{tab:bertscore_length} for details.  Furthermore, our length distribution overlaps significantly with the reference summary lengths, while the BERTScores are consistently higher than the average, at all lengths.  See Figure \ref{fig:bertscore_length}.

\begin{table}
\begin{center}

\begin{tabular}{ c  c c  c  c  c }
\toprule
& Length & BERTScore & Correlation & Slope & Adjusted BERTScore\\
\hline
175b full tree & 719 $\pm$ 321 & 0.182 $\pm$ 0.039 & -0.148 & -3.36e-6 & 0.174 \\ 
175b first subtree & 806 $\pm$ 379  & 0.178 $\pm$ 0.048 & -0.080 & -1.78e-5 & 0.174 \\ 
6b full tree & 655 $\pm$ 308 & 0.125 $\pm$ 0.043 & -0.024 & -1.01e-5 & 0.123 \\
\bottomrule

\end{tabular}

\caption{\label{tab:bertscore_length}Controlling for length with a linear regression does not change BERTScore significantly.  We report correlation and the regression slope.  We target a length of 1167.2 tokens, the average number of tokens in the reference summaries.}
\end{center}
\end{table}

\section{Book summary qualitative findings}
\label{sec:qualitative}

\subsection{Limitations observed by labelers/researchers}
\label{sec:booklimitations}

We chose our task of abstractive summarization of narrative books to be difficult, and our models are still far from human quality.  Here are some of the problems that our labelers reported, roughly in order of frequency and severity.
\begin{itemize}
    
\item The model frequently gets confused between characters, mis-attributing actions.  Interpersonal relationships of the characters were often incorrect and events were wrongly attributed.  Sometimes the name given to the protagonist was a peripheral character, or even the author's name.  This is exacerbated by mis-resolved pronouns and long dialogues, and likely exacerbated by concatenation of summaries.  
\item The model is often unable to pick out the important information, rather than disjointed bits of unimportant stuff.  The “essence” of the story was missing from many summaries.  For example, a summary of A Promised Land never mentioned Obama's presidency.  In books with unique imaginary/speculative elements, the model fails to integrate key world-building details.  This makes some science fiction and fantasy books particularly hard to summarize.
\item Relatedly, the model tends not to abstract away from specific happenings.  For example, judging characters’ mental states, authorial intent, or abstracting a very long chain of events into a single coherent one.
\item The model tends to focus more on earlier material.
\item The model doesn’t handle scene switches or flashbacks well (e.g. the Midnight Library has incursions from different universes, Transcendent Kingdom is non-chronological)
\item Occasionally a quote/excerpt was selected that misrepresented a character or their actions (e.g. a character trying to hide their identity acting like someone else). 
\end{itemize}

\subsection{Preexisting knowledge}

We found that the model was very able to leverage preexisting knowledge from pretraining, often in interesting ways.  

Labelers reported behavior such as the model using the fact that Anakin Skywalker’s daughter is Leia in the Star Wars universe, while it was not mentioned in the passage.  One of the books in full book evaluations was The Ballad of Songbirds and Snakes, a prequel to the previously published Hunger Games trilogy.  A labeler noticed that the model spuriously mentioned characters from the main trilogy who did not appear in the prequel.  Sometimes it uses this data falsely, such as introducing a real world actor’s name into a fictional story in place of a fictional actor.

Another labeler reported that bilingual text was partially translated in the summary, with the model taking “The woman at my mother’s side reached out to touch her—vas a estar bien, she told her before turning to walk back to her car.” and summarizing “The woman accompanying them tells his mother she’ll be ok.”

As a further confirmation of this, we tried summarizing a version of Harry Potter with many characters given replacement names.  Despite this, the model translated “you-know-who” back to Voldemort, despite Voldemort having been given a different name.  

\subsection{Difficulty of summarizing narrative fiction}

Despite the fact that our model was trained on narrative fiction, narrative fiction books seemed to remain more difficult to summarize than other books, due to the reasons outlined in \ref{sec:booklimitations}.

Of the 40 books we chose for the full book evaluations, 6 were nonfiction (see Table \ref{tab:books}).  These 6 books had significantly higher Likert ratings than the fiction books (1st, 2nd, 4th, 5th, 7th, and 11th highest average ratings of model summaries).  Furthermore, the only book which our labelers judged as non-narrative\footnote{Caste was determined to have no plot.} had the 2nd highest Likert ratings.  While this is not strong evidence, it agrees with the qualitative reports from \ref{sec:booklimitations}.

\section{Book summary samples}

We provide a website with examples of project Gutenberg summary trees at \href{https://openaipublic.blob.core.windows.net/recursive-book-summ/website/index.html#/gutenberg}{our website}\footnote{\href{https://openaipublic.blob.core.windows.net/recursive-book-summ/website/index.html\#gutenberg}{https://openaipublic.blob.core.windows.net/recursive-book-summ/website/index.html\#gutenberg}}.  We also provide examples from our test set of books published in 2020.

\subsection{Books used for full book human evaluation}

See \ref{tab:books} for the full list of books we used for the evaluations in Section \ref{sec:fullbookevals}, based on popularity from Goodreads according to \href{https://www.goodreads.com/book/popular_by_date/2020}{this list} at the time we checked.

\subsection{Book samples}

To provide a better understanding of the quality of the summaries generated by our models, we show samples at various Overall Likert scores, ranging from 2 to 6 (Tables \ref{tab:likert2}-\ref{tab:likert6}, for books from the Goodreads test set (that our model has not seen during training time). We select the books at random with the constraint that our 175B first-tree RL policy has one summary that attains the desired Likert score.
For each book, we show the best human-written summary, the 175B RL summary with the desired Likert score, and a random summary from the 175B BC policy at T=0.

\begin{table}
\begin{center}

\begin{small}
\begin{tabular}{ c  c  c  c }

\toprule
\textbf{Title} & \textbf{Author} & \small \textbf{Fiction} & \textbf{Genre} \\ 
\hline

\href{https://www.goodreads.com/book/show/51791252-the-vanishing-half}{The Vanishing Half}	& Brit Bennett & \checkmark & Historical Fiction \\ %
\href{https://www.goodreads.com/book/show/50623864-the-invisible-life-of-addie-larue}{The Invisible Life of Addie LaRue}	& V.E. Schwab & \checkmark & Fantasy \\ %
\href{https://www.goodreads.com/book/show/45046527-american-dirt}{American Dirt}	&  Jeanine Cummins & \checkmark & Contemporary\\ %
\href{https://www.goodreads.com/book/show/51901147-the-ballad-of-songbirds-and-snakes}{The Ballad of Songbirds and Snakes} & Suzanne Collins & \checkmark & Young Adult, Science Fiction \\ 
\href{https://www.goodreads.com/book/show/52129515-untamed}{Untamed} & Glennon Doyle &  & Autobiography \\ 
\href{https://www.goodreads.com/book/show/51933429-the-guest-list}{The Guest List}	& Lucy Foley  & \checkmark & Mystery, Thriller \\ %
\href{https://www.goodreads.com/book/show/44778083-house-of-earth-and-blood}{House of Earth and Blood}	& Sarah J. Maas & \checkmark & Fantasy \\ %
\href{https://www.goodreads.com/book/show/53152636-mexican-gothic}{Mexican Gothic}	& Silvia Moreno-Garcia & \checkmark & Horror \\ %
\href{https://www.goodreads.com/book/show/50093704-in-five-years}{In Five Years}	& Rebecca Serle & \checkmark & Romance \\ %
\href{https://www.goodreads.com/book/show/52867387-beach-read}{Beach Read} & Emily Henry & \checkmark & Romance, Adult Fiction \\ %
\href{https://www.goodreads.com/book/show/44890081-my-dark-vanessa}{My Dark Vanessa} & Kate Elizabeth Russell & \checkmark & Contemporary\\ %
\href{https://www.goodreads.com/book/show/55361205-a-promised-land}{A Promised Land} & Barack Obama &  & Autobiography, Politics \\ 
\href{https://www.goodreads.com/book/show/45754981-the-glass-hotel}{The Glass Hotel} & Emily St. John Mandel & \checkmark & Contemporary, Mystery \\ %
\href{https://www.goodreads.com/book/show/51152447-caste}{Caste
} & Isabel Wilkerson &  & History \\ %
\href{https://www.goodreads.com/book/show/52831200-from-blood-and-ash}{From Blood and Ash} & Jennifer L. Armentrout & 	\checkmark & Fantasy \\ %
\href{https://www.goodreads.com/book/show/45294613-dear-edward}{Dear Edward} & Ann Napolitano & \checkmark & Contemporary \\ %
\href{https://www.goodreads.com/book/show/53287484-midnight-sun}{Midnight Sun} & Stephenie Meyer &  \checkmark & Fantasy, Young Adult \\ %
\href{https://www.goodreads.com/book/show/48570454-transcendent-kingdom}{Transcendent Kingdom}	& Yaa Gyasi &  \checkmark & Contemporary, Literary \\ 
\href{https://www.goodreads.com/book/show/52578297-the-midnight-library}{The Midnight Library} & Matt Haig &  \checkmark & Fantasy \\ %
\href{https://www.goodreads.com/book/show/45047384-the-house-in-the-cerulean-sea}{The House in the Cerulean Sea} & T.J. Klune  & \checkmark & Fantasy \\ %
\href{https://www.goodreads.com/book/show/17699853-chain-of-gold}{Chain of Gold} & Cassandra Clare  &  \checkmark &  Fantasy \\ %
\href{https://www.goodreads.com/book/show/51187948-the-splendid-and-the-vile}{The Splendid and the Vile} & Erik Larson &   & History \\ %
\href{https://www.goodreads.com/book/show/52698452-the-book-of-longings}{The Book of Longings} & Sue Monk Kidd &  \checkmark & Historical Fiction \\ %
\href{https://www.goodreads.com/book/show/50833559-home-before-dark}{Home Before Dark} & Riley Sager &  \checkmark & Horror, Thriller \\ %
\href{https://www.goodreads.com/book/show/52755548-big-summer}{Big Summer} & Jennifer Weiner & \checkmark &  Women's Fiction, Mystery \\ %
\href{https://www.goodreads.com/book/show/45885644-the-sun-down-motel}{The Sun Down Motel} & Simone St. James &  \checkmark & Mystery, Thriller \\ %
\href{https://www.goodreads.com/book/show/35702241-the-shadows-between-us}{The Shadows Between Us} & Tricia Levenseller &  \checkmark & Fantasy \\ %
\href{https://www.goodreads.com/book/show/50214741-the-girl-with-the-louding-voice}{The Girl with the Louding Voice} & Abi Daré &  \checkmark & Contemporary, Cultural \\ %
\href{https://www.goodreads.com/book/show/50548197-a-deadly-education}{A Deadly Education} & Naomi Novik  & \checkmark & Fantasy \\ %
\href{https://www.goodreads.com/book/show/44654627-one-of-us-is-next}{One of Us Is Next} & Karen M. McManus & \checkmark & Mystery, Young Adult \\ %
\href{https://www.goodreads.com/book/show/50892433-one-by-one}{One By One} & Ruth Ware & \checkmark & Mystery, Thriller \\ %
\href{https://www.goodreads.com/book/show/43834909-long-bright-river}{Long Bright River} & Liz Moore  & \checkmark & Mystery \\ %
\href{https://www.goodreads.com/book/show/26082916-ready-player-two}{Ready Player Two} & Ernest Cline & \checkmark & Science Fiction \\ %
\href{https://www.goodreads.com/book/show/51037979-in-an-instant}{In an Instant} & Suzanne Redfearn  & \checkmark &  Contemporary, Young Adult \\ %
\href{https://www.goodreads.com/book/show/52762903-the-book-of-lost-names}{The Book of Lost Names} & Kristin Harmel  & \checkmark & Historical Fiction \\ %
\href{https://www.goodreads.com/book/show/48829708-to-sleep-in-a-sea-of-stars}{To Sleep in a Sea of Stars} & Christopher Paolini  & \checkmark & Science Fiction \\ %
\href{https://www.goodreads.com/book/show/52838315-greenlights}{Greenlights} & Matthew McConaughey  &  & Autobiography \\ %
\href{https://www.goodreads.com/book/show/50202953-piranesi}{Piranesi} & Susanna Clarke & \checkmark & Fantasy \\ %
\href{https://www.goodreads.com/book/show/50358031-leave-the-world-behind}{Leave the World Behind} & Rumaan Alam  & \checkmark & Thriller, Mystery \\ %
\href{https://www.goodreads.com/book/show/50088631-hidden-valley-road}{Hidden Valley Road
} & Robert Kolker  &  & Psychology, Science \\ %

\bottomrule

\end{tabular} 

\end{small}
\caption{\label{tab:books}Books used for final evaluations.  All books were published in 2020, and thus do not appear in our pretraining data. 
Genres are determined by taking the top two Goodreads user-labeled genres; ignoring "Fiction", "Nonfiction", "Audiobook", and near-duplicates; and excluding the second if it has less than half the count of the first.
}
\end{center}
\end{table}


\begin{table}[]
    \centering
    \small
    \begin{tabular}{>{\arraybackslash}m{12cm} l}
    \toprule 
    \textbf{Summaries of ``In an Instant'' by Suzanne Redfearn (Likert=2)} & \textbf{Ratings}\\ \hline
        \multicolumn{1}{c}{\textbf{Best human-written  summary}} \\
        Finn Miller is a ghost watching over her family and friends –the survivors of the car crash that killed her. She watches them in the moments after she died, and after Ann and Kyle manage to call for rescue.
 
Her parents, Jack and Ann, had been on the verge of divorce. Despite losing two of their four children and Ann’s cheating (with Bob), they rediscover their love for one another and start working together as a couple.
Her best friend, Maureen, loyal to Finn’s family, investigates the truth about what happened after the accident. Maureen bonds with Kyle (who they met during the accident), when she asks him about what he remembers about the crash. She and Kyle fall in love much to Finn’s delight.
 
Chloe (Finn’s sister) and  Vance (Chloe’s boyfriend), are lost and separated in the snow. They are eventually rescued but Chloe realizes she needs to move on from the relationship –she finds a new future (and love life) volunteering for an animal shelter. Vance moves on too, now sober from drugs and more responsible, thanks to Jack’s help.
 
Her brother Oz, has emotional growth issues and dies in the cold after being tricked by Bob (a family friend they treat as an uncle). When the sheriff learns about this after Maureen’s investigation, Bob is arrested for negligent homicide. While not obvious to Maureen, the Sheriff understood that since Bob was the adult, Bob had a responsibility to keep Maureen and Oz safe.
        & \multirow{5}{*}{\shortstack[l]{
Overall: 6.3\\
Accuracy: 7.0\\
Coverage: 6.7\\
Coherence: 5.7\\
Abstraction: 3.0}}
        
        \\ \hline
        
        \multicolumn{1}{c}{\textbf{175B RL summary attaining an Overall Likert of 2.0}} \\ 
        Aubrey's latest car accident has damaged her mom's new car, and her family is tense. Mo is excited to go skiing for the first time, but the camper crashes, killing Mo. Her family is stranded in a blizzard miles from help.

Aubrey tries to be the ray of light her family needs, despite her own feelings of misery.

Vance has been searching for his brother Oz for five days, but it's clear that Oz's body is long gone.

At Aubrey's wedding, Mo notices her mom standing alone on the sidelines and sends Kyle over to ask her to dance. As they begin to dance, the world lightens and the edges begin to glow.
& \vspace{4mm}\multirow{5}{*}{\shortstack[l]{Overall: 2.0\\
Accuracy: 1.5\\
Coverage: 2.0\\
Coherence: 3.0\\
Abstraction: 4.5}}
        \\ \hline 
       \multicolumn{1}{c}{\textbf{Random 175B BC summary}}\\  
        Finn's sister, Aubrey, is getting married. She is upset about her wedding dress. Finn calls Mrs. Kinsell, the woman who is going to make her wedding dress, and asks her to make a different dress.

A blizzard traps a group of people in a camper. Finn's dad is unconscious and his leg is broken. Mo and Kyle go to get help. Mo's mom has mild frostbite on her fingers, several of her toes, and patches of her calves where snow and ice.

Vance drives Mo's dad and himself to the cabin. He says that he's leaving. Mo's dad says that he's kidnapping him. He says that he needs to turn him back into the arrogant punk he used to be so Chloe can realize what an asshole he is.

Chloe says that she wrote down what she knows about the accident. She says that her mom won't talk about it. Chloe says that she thinks the part of her mom's story that they don't know is the part that has her racing through the streets like a madwoman and pretending she only ever had two children instead of four.

Mo drives up the winding road toward Big Bear. Burns shows her a newspaper article. The headline reads, Five Rescued from Crash after Night in the Snow. The photo shows Uncle Bob hobbling from a Forest Service helicopter with his arms slung over two rescue workers. Natalie is almost out of view but visible enough to see a bright-purple glove sticking out from the sleeve of her long down coat. Burns asks Mo if Oz was dangerous. She says no. Mo realizes that no one knows what really happened.
        & \multirow{5}{*}{\shortstack[l]{Overall: 2.5\\
Accuracy: 3.0\\
Coverage: 3.0\\
Coherence: 3.5\\
Abstraction: 4.0}}
        \\
    \bottomrule
    \end{tabular}
    \caption{Summaries of ``In an Instant'' written by our models (175B RL policy and 175B behavior cloning policy) and a human. Sample selected at random with the constraint that a 175B RL summary achieved an Overall Likert score of 2.0. See \href{https://openaipublic.blob.core.windows.net/recursive-book-summ/website/index.html}{our website} for all human and model summaries. }
    \label{tab:likert2}
\end{table}

\begin{table}[]
    \centering
    \small
    \begin{tabular}{>{\arraybackslash}m{12cm} l}
    \toprule 
    \textbf{Summaries of ``Transcendent Kingdom'' by Yaa Gyasi (Likert=3)} & \textbf{Ratings}\\ \hline
        \multicolumn{1}{c}{\textbf{Best human-written  summary}} \\
        $[$The book is a non-chronological reflection on stories from Gifty’s life, tied together by her philosophical thoughts.$]$
Gifty’s parents grow up in Ghana, marrying at 30. Gifty’s mother decides they should emigrate to America to give their first child, Nana, room to grow. Gifty's father disagrees but capitulates due to his laid-back nature.
In Alabama, Gifty’s mother works long hours as a caregiver for the elderly and endures their racist abuse.  As Gifty grows up her parents are emotionally distant and she devotes herself to school and her evangelical religion.
Her father, alienated in America, abandons them and returns to Ghana.
Nana is a gifted basketball player in high school. After an injury, he becomes addicted to opioid painkillers. He dies of an overdose after several failed attempts to get clean.
Nana’s death destroys eleven-year-old Gifty’s faith and sends her mother into a severe depressive episode, leaving her unresponsive in bed. Gifty finds her mother after a failed suicide attempt and spends a summer with her aunt in Ghana while she recovers.
Functional but still depressed, her mother remains distant as Gifty excels in school.
Gifty studies neuroscience at Harvard and then Stanford, focusing on reward regulation mechanisms in mice. She tells herself she chose the field for the challenge and prestige but inwardly hopes to understand and potentially cure addiction and depression. She pushes away any close relationships and focuses single-mindedly on her Ph.D.
When her mother suffers another depressive episode, Gifty brings her to her small apartment to care for her. For the first time, she tells her colleagues about her past. Together they help Gifty's mother begin a slow recovery.
        & \multirow{5}{*}{\shortstack[l]{
Overall: 7.0\\
Accuracy: 7.0\\
Coverage: 7.0\\
Coherence: 7.0\\
Abstraction: 7.0}}
        
        \\ \hline
        
        \multicolumn{1}{c}{\textbf{175B RL summary attaining an Overall Likert of 3.0}} \\ 
        Gifty is a PhD candidate in neuroscience at Stanford University. Her brother Buzz died of a heroin overdose three months after their parents immigrated to America. Her father cowardly left the family, and her mother has to take on two jobs to make ends meet.

Years later, Gifty returns home to find her mother cooking in the kitchen. After Nana's basketball career takes off, he's injured during a game and taken to the hospital. He becomes hooked on OxyContin and dies of an overdose of heroin in a Starbucks parking lot.

After two weeks at the UAB psychiatric hospital, Gifty's mother returns home. She seems to be recovering. After six months of dating, Raymond accuses Gifty of being crazy and leaves. Gifty comes home to find her mother missing. She frantically searches the apartment complex, calling Katherine for help. Her mother tells her not to be afraid, that God is with her wherever she goes.
& \multirow{1}{*}{\shortstack[l]{Overall: 3.0\\
Accuracy: 4.0\\
Coverage: 3.0\\
Coherence: 3.0\\
Abstraction: 3.5}}
        \\ \hline 
       \multicolumn{1}{c}{\textbf{Random 175B BC summary}}\\  
        The narrator's mother is in a bed for months at a time. The narrator is sent to Ghana to wait her out. The narrator's aunt points out a crazy person in the market.

The narrator's mother and the Chin Chin Man fight. The Chin Chin Man doesn't want to go to the funeral, but the narrator's mother insists. The Chin Chin Man pulls over. The narrator's mother says that they have to pray. The Chin Chin Man says that they can't wait. The narrator's mother says that they must pray now. The narrator's mother says that they must pray. The narrator's mother and the Chin Chin Man fight. The Chin Chin Man doesn't want to go to the funeral, but the narrator's mother insists. The Chin Chin Man pulls over. The narrator's mother asks Nana to try harder. She says that he says that it feels like everything inside his head empties out. She says that she feels something come over her. She raises her hand. Pastor John calls her to the altar. She feels the pressure of his hand like a beam of light from God himself. She starts volunteering at her church. She says that Nana is healing up nicely, but he is still being cautious with his right side. She says that her mother finds more hidden in his light fixture. She says that she is scared for Nana. She says that she is scared of Nana and the death rattle that none of them wanted to acknowledge they were listening for. She says that she knows that it will come. She says that she doesn't know how her mother convinced Nana to accompany them to the First Assemblies one Sunday.
        & \multirow{5}{*}{\shortstack[l]{Overall: 1.5\\
Accuracy: 3.0\\
Coverage: 2.0\\
Coherence: 1.0\\
Abstraction: 3.0}}
        \\
    \bottomrule
    \end{tabular}
    \caption{Summaries of ``Transcendent Kingdom'' written by our models (175B RL policy and 175B behavior cloning policy) and a human. Sample selected at random with the constraint that a 175B RL summary achieved an Overall Likert score of 3.0. See \href{https://openaipublic.blob.core.windows.net/recursive-book-summ/website/index.html}{our website} for all human and model summaries. }
    \label{tab:likert3}
\end{table}

\begin{table}[]
    \centering
    \small
    \begin{tabular}{>{\arraybackslash}m{12cm} l}
    \toprule 
    \textbf{Summaries of ``The Vanishing Half'' by Brit Bennett (Likert=4)} & \textbf{Ratings}\\ \hline
        \multicolumn{1}{c}{\textbf{Best human-written  summary}} \\
        Vignes twins Desiree and Stella (light-skinned black women) leave their town and their mother Adele. In New Orleans, Stella leaves Desiree. Desiree finds work at D.C., where he marries Sam and has a daughter (Jude). Because he beats her, Desiree returns to Mallard with Jude. Adele is surprised by Jude’s dark skin.
 
Early, a hunter, is tasked to search for Desiree. When he finds out Sam beats her, he tells Sam he cannot track her. The two become a couple. 

Jude studies at UCLA and gets into a relationship with Reese, a transsexual.
 
Stella (now Stella Sanders), who hides that she’s part-black, is married to a white man (Blake), and has a child (Kennedy). She suffers from depression after Loretta (a black woman she befriended) and her family leave. She continues her studies and becomes a Statistics Instructor.
 
Jude meets Kennedy (a theater actress) by chance and finds out she’s Stella's daughter. Jude sees Stella and approaches her but Stella seems repulsed by her color. Jude and Kennedy meet by chance again in New York, when Jude accompanies Reese for his chest removal surgery. 
 
In 1981, the twins reunite. Stella (who did not say goodbye) gives Early her wedding ring, telling him to sell it and to take care of Adele (who has Alzheimer’s). 
 
Jude (in medical school) still contacts Kennedy sometimes. Her mom calls to say Adele has died. After the funeral, Desiree and Early sell the house and move to Houston.
        & \multirow{5}{*}{\shortstack[l]{
Overall: 5.7\\
Accuracy: 6.0\\
Coverage: 5.3\\
Coherence: 5.7\\
Abstraction: 4.7}}
        
        \\ \hline
        
        \multicolumn{1}{c}{\textbf{175B RL summary attaining an Overall Likert of 4.0}} \\ 
        In 1968, the town of Mallard has been searching for the Vignes twins, Stella and Desiree, for 14 years since they disappeared from their bedroom one morning after the Founder's Day dance.

Desiree's daughter, Jude, has been bullied throughout school because of her dark skin. Her mother has lied to her about her father, who hasn't been in contact with them in 10 years.

Jude has always wanted to be different, but none of the potions her grandmother makes work. She's been spending every weekend with Reese, so inseparable that people begin to ask for one if they see the other.

Stella has been married to Blake for 8 years and has lied to him about her family, telling him her parents died in an accident. She enjoys her comfortable life with her daughter, Kennedy.

In 1988, Stella returns to her hometown after years away. She explains that she'd met her former boss, Mr. Sanders, and ran off with him. Stella has a daughter, Kennedy, who's an actor in New York. Stella begs Desiree not to take her daughter away from her, explaining that she hasn't been a good mother.

Jude enjoys working in the lab, fascinated by the challenge of searching for tiny nerves. She's been in love with Reese since she was 18 and doesn't think she can go back to her hometown without him.
& \multirow{1}{*}{\shortstack[l]{Overall: 4.0\\
Accuracy: 5.0\\
Coverage: 4.0\\
Coherence: 4.5\\
Abstraction: 3.0}}
        \\ \hline 
       \multicolumn{1}{c}{\textbf{Random 175B BC summary}}\\  
        Lou LeBon runs to the diner to break the news that Desiree Vignes, one of the lost twins, has returned to Mallard. She's holding the hand of a girl, 7 or 8, and black as tar. The townspeople wonder if the girl is Desiree's. The twins ran away to New Orleans, but they never returned. Stella became white and married a white man, while Desiree married a black man and had a black daughter.

Early Jones is good at getting lost. He's been running jobs for Ceel for seven years. Ceel is a bail bondsman. He's looking for a new bounty hunter. He notices Early. He says that a man wants his wife found. She ran off with his kid.

Early and Desiree search for Stella. They go to the Maison Blanche building. A white security guard helps her. She says that she's looking for an old friend who used to work in marketing. The secretary says that she doesn't have an address for her. She says that she has a forwarding address for her in Boston. Early says that it's a start. They go to the Surly Goat. They kiss.

Stella Vignes has been Stella Vignes in New Orleans and Boston, but Early can't find a marriage license for a Stella Vignes in any place he knows she's been. He's been on the trail for years. He's been caught between Desiree and Stella.

Reese Carter becomes Therese Anne Carter. He cuts his hair in a truck stop bathroom with a stolen hunting knife. He becomes a professional photographer. He spends his weekends shooting abandoned buildings

Stella Vignes is a 21-year-old secretary. She's married to Blake Sanders, a 28-year-old marketing department at Maison Blanche. Stella's
        & \multirow{5}{*}{\shortstack[l]{Overall: 3.5\\
Accuracy: 5.0\\
Coverage: 2.5\\
Coherence: 4.5\\
Abstraction: 2.5}}
        \\
    \bottomrule
    \end{tabular}
    \caption{Summaries of ``The Vanishing Half'' written by our models (175B RL policy and 175B behavior cloning policy) and a human. Sample selected at random with the constraint that a 175B RL summary achieved an Overall Likert score of 4.0. See \href{https://openaipublic.blob.core.windows.net/recursive-book-summ/website/index.html}{our website} for all human and model summaries. }
    \label{tab:likert4}
\end{table}

\begin{table}[]
    \centering
    \small
    \begin{tabular}{>{\arraybackslash}m{12cm} l}
    \toprule 
    \textbf{Summaries of ``Big Summer'' by Jennifer Weiner (Likert=5)} & \textbf{Ratings}\\ \hline
        \multicolumn{1}{c}{\textbf{Best human-written  summary}} \\
        Daphne Berg, a plus-size influencer, agrees to promote Leela Thakoon’s clothing line. She thinks about how she decided to accept her body after being deeply hurt by her rich and beautiful friend Drue. However, Daphne has recently agreed to go to Drue’s wedding. 
The night before the wedding, Daphne meets Nick and they make love, but in the morning, Daphne finds Drue dead in the hot tub. Daphne and Darshi start investigating. They find Nick and Emma Vincent’s mother. Emma was arrested for Drue’s murder, but her mother explains that Drue’s father is both Emma’s and Nick’s father too. Nick is shocked, since his mother was murdered when he was a child and he didn’t know anything. 
Daphne finds out that Drue’s marriage was a ruse so Drue could get money to save her father’s business and finally get his approval. They also find Drue’s secret boyfriend, Aditya Acharya. Aditya tells them that Drue cheated on her SATs, while the girl who took the tests for her was expelled. Daphne realizes that that girl is Leela and that she killed Drue. 
Daphne goes to her apartment, where Leela is waiting for her with a gun. Daphne manages to overpower her.
Daphne apologizes to Darshi for always chasing after Drue, while Darshi was the one who was always there for her. She and Nick make love and she feels grateful for all the people around her that love her, which is something Drue never had.
        & \multirow{5}{*}{\shortstack[l]{
Overall: 7.0\\
Accuracy: 7.0\\
Coverage: 7.0\\
Coherence: 6.7\\
Abstraction: 4.0}}
        
        \\ \hline
        
        \multicolumn{1}{c}{\textbf{175B RL summary attaining an Overall Likert of 5.0}} \\ 
        Daphne is thrilled to be offered a job as the exclusive face and figure of Leef Fashion. After a viral video of her fighting a guy who called her fat goes viral, her parents fuss over her, treating her like an invalid.

At Drue and Stuart's engagement party, Drue tells Daphne that she's getting married tomorrow. Later, Nick tells Daphne that he'd seen Drue's fiancé, Stuart, kissing another girl on the beach.

The next morning, Daphne wakes up to find Drue's body floating facedown in the hot tub. Daphne is questioned by Detective Ryan McMichaels about the events of the previous night. She tells him everything she knows, insisting that she had no reason to want to hurt Drue.

After Drue's funeral, Detective McMichaels tells Daphne that Emma is no longer a suspect in Drue's murder. He warns that Nick could have had a motive to kill Drue if he's her half-brother. Daphne and Nick research Drue's past and find a scandal at Croft School where Drue had gotten a girl to take the SATs for her. They suspect Kamon Charoenthammawat, who'd changed her name to Leela Thakoon, is the killer. Leela shows up at Darshi's apartment and points a gun at her, threatening to kill her if she doesn't take her own life.
& \multirow{1}{*}{\shortstack[l]{Overall: 5.0\\
Accuracy: 4.5\\
Coverage: 5.0\\
Coherence: 5.5\\
Abstraction: 5.0}}
        \\ \hline 
       \multicolumn{1}{c}{\textbf{Random 175B BC summary}}\\  
        Christina is alone in the summer cottage on the outer Cape. Her mother is dead, her sisters and brother are strangers, and her father is furious when he learns the reason why she wants the cottage. She tells him she's pregnant and that she's keeping the baby. He relents.

Christina gives birth in Boston and then takes herself and her baby to the cottage. She's worried Aidan will fuss on the ride up, but he sits like a wise old owl. She loves the cottage, which isn't grand but is perfect for her. She makes friends and acquires what she needs. She makes a life for herself and her son.

Christina is hired by Leela to work with her. She's excited about the job, but she's worried about the increased attention. She's worried about the trolls. She's worried about the people who are revolted by her body. She's worried about the people who are disgusted by her body. She's worried about the people who are concerned about her body. She goes live on Instagram. She tells her fans that she's got some good news. She tells them to go to her YouTube channel and watch the first video she ever posted. She tells them that she's stopped dieting and started being herself.

Christina goes out with her friends. Lake asks Christina if she wants another drink. She says yes. Lake goes to get her a drink. Lake and Drue see Christina. Lake takes her hand and leads her to the dance floor. They dance to a slow song. Lake puts his hands on her. She thinks about Drue kissing Christina and tells her to post the pictures. She posts them and then Nick leads her to the buffet. They eat and then he starts to talk about himself. He tells her about the time he and his cousins took their grandmother to play bingo in Provincet
        & \multirow{5}{*}{\shortstack[l]{Overall: 3.0\\
Accuracy: 2.5\\
Coverage: 2.5\\
Coherence: 4.5\\
Abstraction: 3.5}}
        \\
    \bottomrule
    \end{tabular}
    \caption{Summaries of ``Big Summer'' written by our models (175B RL policy and 175B behavior cloning policy) and a human. Sample selected at random with the constraint that a 175B RL summary achieved an Overall Likert score of 5.0. See \href{https://openaipublic.blob.core.windows.net/recursive-book-summ/website/index.html}{our website} for all human and model summaries. }
    \label{tab:likert5}
\end{table}

\begin{table}[]
    \centering
    \small
    \begin{tabular}{>{\arraybackslash}m{12cm} l}
    \toprule 
    \textbf{Summaries of ``Dear Edward'' by Ann Napolitano (Likert=6)} & \textbf{Ratings}\\ \hline
        \multicolumn{1}{c}{\textbf{Best human-written  summary}} \\
        On June 12, 2013, a plane travelling from NY to LA crashes due to the co-pilot’s mistake. There is only one survivor, twelve-year-old Edward Adler. The book switches between the events on the plane and Edward’s life after the crash. 

Besides Edward’s family, other passengers include Linda Stollen, a young woman travelling to meet her boyfriend Gary, Crispin Cox, a bitter rich old man fighting cancer, Florida, a free-spirited woman who can remember her past lives, Benjamin Stillman, an injured black soldier, and Mark Lassio, a successful Wall Street man.  

After the crash, Edward is adopted by his aunt Lacey and his uncle John. They are very caring and supportive, but their marriage is faltering. Edward is depressed and traumatized, but finds comfort in his neighbor Shay, who is his age. 

Two years after the crash, Edward finds all the documents John has kept about the crash and the passengers. He also finds letters from passengers’ relatives addressed to him. He and Shay read them and categorize them. When he reads a letter from Mahira, his brother’s girlfriend no one knew about, he goes to meet her. Jax Lassio, Mark Lassio’s brother, leaves Edward all of Mark’s money, and Edward decides to use it to anonymously help the people around him and the passengers’ loved ones. 

In June 2019, Edward and Shay visit the memorial site in Colorado and Edward feels he has come full circle. He and Shay kiss.
        & \multirow{5}{*}{\shortstack[l]{
Overall: 6.7\\
Accuracy: 7.0\\
Coverage: 6.3\\
Coherence: 7.0\\
Abstraction: 5.7}}
        
        \\ \hline
        
        \multicolumn{1}{c}{\textbf{175B RL summary attaining an Overall Likert of 6.0}} \\ 
        After a plane crash in Colorado, Edward is the only survivor. He's severely injured and is taken to live with his aunt and uncle. He's unable to answer any questions about how he's feeling.

A year later, Edward is awarded \$5 million in compensation for the crash, but he doesn't want the money.

Edward's health is given the all clear, which means he has to attend summer camp with Shay.

Edward receives a letter from Mark informing him that he's left him \$7.3 million in his will. He decides to use the money to give to others.

Edward and Shay visit the site of the plane crash, and he feels nauseous as he recalls the memories of the crash. He thinks about how the loss of his brother Jordan will always be painful.
& \vspace{4mm}\multirow{1}{*}{\shortstack[l]{Overall: 6.0\\
Accuracy: 6.0\\
Coverage: 5.0\\
Coherence: 6.0\\
Abstraction: 4.5}}
        \\ \hline 
       \multicolumn{1}{c}{\textbf{Random 175B BC summary}}\\  
        Edward is the only survivor of a plane crash. He is famous because he is the boy who was pulled from the wreckage and survived. He is famous because he is the boy who was pulled from the wreckage and survived. He is famous because he is the boy who was pulled from the wreckage and survived. He is famous because he is the boy who was pulled from the wreckage and survived. He is famous because he is the boy who was pulled from the wreckage and survived. He is famous because he is the boy who was pulled from the wreckage and survived. He is famous because he is the boy who was pulled from the wreckage and survived. He is famous because he is the boy who was pulled from the wreckage and survived.
        & \vspace{4mm}\multirow{2}{*}{\shortstack[l]{Overall: 1.0\\
Accuracy: 2.5\\
Coverage: 1.0\\
Coherence: 2.0\\
Abstraction: 1.5}}
        \\
    \bottomrule
    \end{tabular}
    \caption{Summaries of ``Dear Edward'' written by our models (175B RL policy and 175B behavior cloning policy) and a human. Sample selected at random with the constraint that a 175B RL summary achieved an Overall Likert score of 6.0. See \href{https://openaipublic.blob.core.windows.net/recursive-book-summ/website/index.html}{our website} for all human and model summaries. }
    \label{tab:likert6}
\end{table}

\begin{table}[]
    \centering
    \begin{tabular}{>{\arraybackslash}m{14cm}}
    \toprule 
    \textbf{175B RL summary of ``Harry Potter and the Philosopher's Stone'' by J.K.Rowling} \\ \hline
        
Harry Potter lives with his aunt and uncle since his parents were killed in a car crash when he was a baby. On his 11th birthday, Hagrid, the Keeper of Keys and Grounds at Hogwarts, informs Harry that he's a wizard and has been accepted to Hogwarts.

Harry quickly becomes a natural at Quidditch and is appointed the new Gryffindor Seeker. During a Quidditch match, Harry follows Snape into the forest and sees him talking to Quirrell. He realizes Snape is after the Sorcerer's Stone.

Harry, Ron, and Hermione decide to steal the Stone themselves. They manage to get past several obstacles and reach the final chamber, where Quirrell appears and reveals that he let the troll into the school to kill Harry. Harry manages to escape and wakes up in the hospital wing. Dumbledore explains that the Stone has been destroyed. \\
    \bottomrule
    \end{tabular}
    \caption{A bonus summary of ``Harry Potter and the Philosopher's Stone'' by our 175B RL model. }
    \label{tab:hp}
\end{table}

\end{document}